\pdfoutput=1

\documentclass[11pt]{article}
\usepackage[table,dvipsnames]{xcolor}
\usepackage[final]{acl}
\usepackage{times}
\usepackage{latexsym}
\usepackage{graphicx}
\usepackage{subcaption}
\usepackage{amsmath}
\usepackage{siunitx}
\usepackage{multirow}
\usepackage{threeparttable}
\usepackage{bm}
\usepackage{pdfpages}

\usepackage[T1]{fontenc}
\usepackage{paralist}

\usepackage[utf8]{inputenc}
\usepackage{listings}

\usepackage{tikz}
\usetikzlibrary{arrows.meta, positioning, calc}

\usepackage{microtype}

\usepackage{inconsolata}
\usepackage{multirow}
\usepackage{makecell}

\usepackage{comment}
\usepackage{subcaption}
\usepackage{booktabs}
\usepackage{array}
\usepackage{pgfplots}
\pgfplotsset{compat=1.18}

\usepackage{adjustbox}
\usetikzlibrary{patterns}
\usepackage{longtable}      %
\usepackage{colortbl}       %
\usepackage{hyperref}       %
\usepackage{cleveref}       %
\usepackage{xfp}
\usepackage{fvextra}

\newcommand{\barwidth}{7} %

\newcommand{\barheight}{7.5pt} %
\newcommand{\percentscale}{100} %

\definecolor{myblue}{RGB}{50,188,221}

\def\pcb#1{%
   {\color{myblue}\rule{\fpeval{#1/\percentscale*\barwidth} cm}{\barheight}} #1
}
\usepackage{mdframed}
\usepackage{xcolor}

\newenvironment{csquote}{
    \begin{mdframed}[
        linewidth=1pt,
        linecolor=gray!40,
        backgroundcolor=gray!5,
        leftmargin=1em,
        rightmargin=1em,
        innertopmargin=8pt,
        innerbottommargin=8pt
    ]
}{
    \end{mdframed}
}

\newcommand{\interviewq}[2]{
    \textbf{Q#1:} #2\\[0.5em]
}

\title{iNews: A Multimodal Dataset for Modeling Personalized Affective Responses to News}

\author{Tiancheng Hu\\
  University of Cambridge\\
  \texttt{th656@cam.ac.uk} \\
    \raisebox{-0.2\height}{\includegraphics[height=1em]{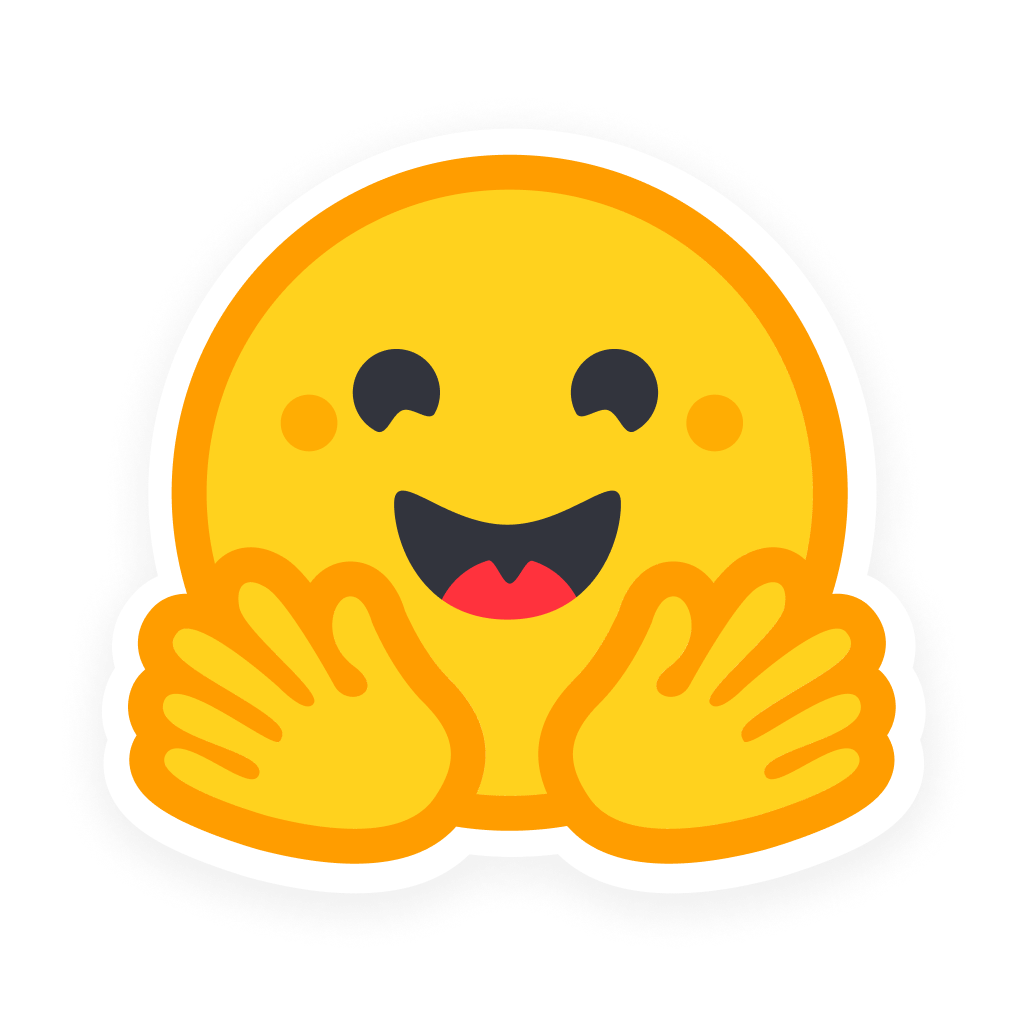}} \href{https://huggingface.co/datasets/pitehu/inews}{Data} \hspace{0.3cm}\\
  \And
  Nigel Collier \\
  University of Cambridge\\
  \texttt{nhc30@cam.ac.uk} \\
    \raisebox{-0.2\height}{\includegraphics[height=1em]{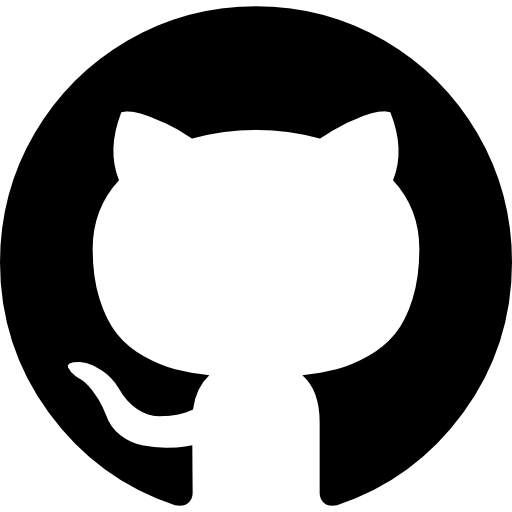}} \href{https://github.com/pitehu/inews/}{Code}
    }

\begin{document}
\maketitle
\newcommand{\barchart}[2]{%
  \begin{tikzpicture}[baseline]
    \fill[orange!80] (0,0) rectangle (#1*\textwidth/100-0.5pt, 6pt);
    \node[anchor=west] at (0, 3pt) {#2};
  \end{tikzpicture}%
}
\begin{abstract}
Understanding how individuals perceive and react to information is fundamental for advancing social and behavioral sciences and developing human-centered AI systems. Current approaches often lack the granular data needed to model these personalized responses, relying instead on aggregated labels that obscure the rich variability driven by individual differences. We introduce iNews, a novel large-scale dataset specifically designed to facilitate the modeling of personalized affective responses to news content. Our dataset comprises annotations from 291 demographically diverse UK participants across 2,899 multimodal Facebook news posts from major UK outlets, with an average of 5.18 annotators per sample. For each post, annotators provide multifaceted labels including valence, arousal, dominance, discrete emotions, content relevance judgments, sharing likelihood, and modality importance ratings. Crucially, we collect comprehensive annotator persona information covering demographics, personality, media trust, and consumption patterns, which explain 15.2\% of annotation variance - substantially higher than existing NLP datasets. Incorporating this information yields a 7\% accuracy gain in zero-shot prediction and remains beneficial even with 32-shot in-context learning. iNews opens new possibilities for research in LLM personalization, subjectivity, affective computing, and human behavior simulation.
\end{abstract}

\section{Introduction}
\begin{figure*}[!htbp]
    \centering
    \includegraphics[width=1\linewidth]{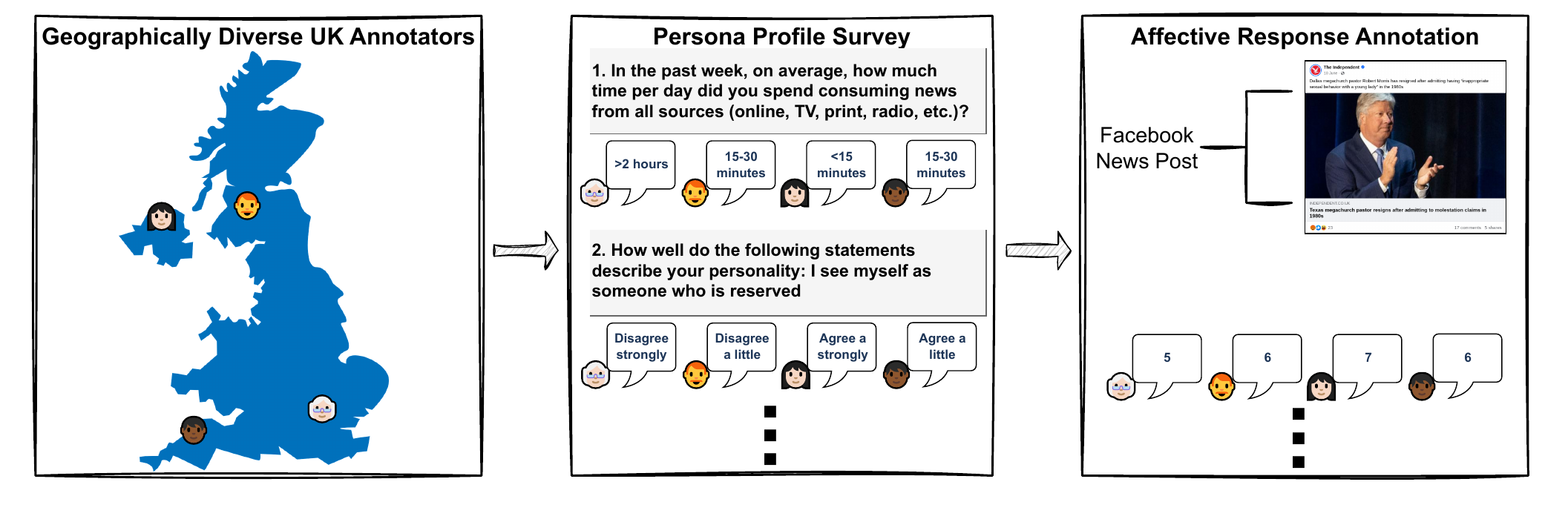}
    \caption{Overview of the data collection process.  The process involves three main stages: (1) we recruit demographically diverse UK annotators; (2) annotators complete a persona profile survey capturing demographics, ideology, news consumption, cognitive traits, personality, and emotional characteristics; and (3) annotators provide affective response annotations for Facebook news posts, including valence, arousal, dominance, discrete emotions, modality influence, personal relevance, and sharing likelihood.}
    \label{fig:data_collection}
\end{figure*}

Understanding and predicting individual human behavior represents a central challenge across social and behavioral sciences, with applications ranging from public health interventions to economic policy design. Researchers have long recognized that individual characteristics, including personality traits, demographics, cultural backgrounds, and personal experiences, fundamentally shape how people respond to identical stimuli~\cite{kring1998sex, costa2008revised, mesquita1992cultural}. However, computational approaches to modeling human responses in natural language processing (NLP) have largely overlooked this rich individual variability, instead relying on aggregated labels that obscure the person-specific patterns that drive real-world behavioral phenomena~\cite{plank-2022-problem, Cabitza_Campagner_Basile_2023}.

This limitation becomes particularly problematic in NLP applications that aim to understand or predict human responses to textual content. Current approaches typically rely on generic group-level models that ignore individual differences in how people interpret and emotionally respond to identical texts. The emergence of large language models (LLMs), to which psychological theories are increasingly applied for analysis and development~\cite{zhu2024conformity,liu2025mind, hu2025generative}, presents new opportunities for modeling these individual-level differences, but requires datasets that capture both behavioral outcomes and the personal characteristics that drive them.

Affective responses to news content represent an ideal testing ground for benchmarking and developing individual-level behavioral simulation models. This domain offers several key advantages: (1) emotional reactions are observable behavioral outcomes with established measurement frameworks~\cite{mehrabian1974approach, bradley1994measuring, Ekman01051992}, (2) individual differences in affective news processing are well-documented in psychology and media studies~\cite{oliver2002individual,valkenburg2013differential,soroka2019cross}, (3) the task involves real-world stimuli that people encounter daily. Furthermore, news consumption behavior has direct societal relevance for understanding information processing, media effects, and the development of responsible information systems.

We introduce iNews, a novel large-scale dataset specifically designed to capture the inherent subjectivity of affective responses to real-world news content (overview in Figure \ref{fig:data_collection}). Our dataset comprises fine-grained affective responses from 291 annotators to 2,899 Facebook posts from leading UK media outlets. The annotations include: valence-arousal-dominance (VAD) ratings \cite{mehrabian1974approach, bradley1994measuring}, Ekman's basic emotions \cite{Ekman01051992}, perceived post relevance, modality importance, and sharing likelihood. In addition, we collect a comprehensive set of annotator characteristics\footnote{Throughout this work, we also use the term persona information to refer to the same concept.} (e.g. demographics, personality, media consumption habits), drawing upon insights from the differential media effects literature. %

Our regression analysis confirms that annotator characteristics explain a substantial portion of the annotation variance (15.2\% - higher than observed in any NLP dataset to date), highlighting the importance of incorporating individual differences when modeling subjective phenomena like affect. Furthermore, through an open-ended questionnaire with a subset of annotators ($N=20$), we identify nuanced patterns in how individuals experience and articulate their emotional reactions to news content, extending beyond the scope of our structured annotations and survey.

In a case study demonstrating the practical value of this rich persona information, we show that incorporating annotator characteristics can improve LLM predictions of individual-level affective responses by up to 7\% in accuracy in zero-shot settings, although overall accuracy remains relatively modest (around 40\%). When comparing input modalities (image vs. text), we find that image inputs typically outperform text in zero-shot scenarios but this advantage diminishes in few-shot settings. In the few-shot setting, we observe the ``early ascent phenomenon''~\cite{lin2024dual, agarwal2024manyshot}, where performance initially dips below zero-shot levels with very few examples before improving as the number of shots increases. We ultimately reach 44.4\% accuracy at 32-shot. Even at this level, incorporating persona information yields additional performance gains, suggesting that persona-based and example-based approaches provide complementary signals for modeling individual differences.

The iNews dataset benefits a wide range of research areas: affective computing researchers modeling emotion recognition while accounting for individual differences; LLM developers advancing personalization and subjective phenomena handling; human behavior simulation researchers modeling individual-level information processing; social computing scholars investigating demographic effects in content presentation; and AI alignment researchers studying preference diversity across human populations.

\section{Related Work}
\subsection{News, Emotion, and Individual Differences}
The interplay between news content, emotional responses, and downstream cognitive and behavioral effects is often an area of focus in communication and psychology.  Prior research establishes that news often exhibits a negativity bias, eliciting negative emotions and heightened arousal in readers~\cite{soroka2019cross}. However, individual responses vary considerably based on demographic factors, pre-existing political attitudes and identities, personality traits, and other individual and group-level characteristics~\cite{oliver2002individual,valkenburg2013differential,soroka2019cross}. This heterogeneity carries significance beyond immediate emotional experiences, fundamentally influencing information processing and behavior. Emotions provide evaluative feedback, impacting veracity judgments~\cite{Martel2020} and shaping reasoning and decision-making~\cite{marcus2000affective, storbeck2008affective}.

Existing research on the affective dimension of news perception predominantly focuses on the emotional tone of the news content itself, rather than the induced emotional responses of individual readers~\cite{de2020news}. Much of this work relies on aggregate-level analysis, obscuring individual-level variation. Our work addresses these limitations by redirecting attention to fine-grained reader responses. We present a large-scale dataset designed to capture and analyze the spectrum of individual affective responses to news headlines, facilitating a more nuanced understanding of the relationship between news, emotion, and individual differences.

\subsection{Emotion Detection in NLP}

Emotion detection has been a long-standing focus within NLP~\cite{strapparava2007semeval,plaza-del-arco-etal-2024-emotion}. Recent years have seen a large number of valuable resources on the task (see~\citet{demszky-etal-2020-goemotions, oberlander2020goodnewseveryone, plaza-del-arco-etal-2020-emoevent} for a overview). These efforts have significantly advanced the field, leading to more accurate and robust emotion detection systems. 

However, most existing datasets rely on aggregated ``gold labels'', overlooking the inherent subjectivity and variation in human emotional perception~\cite{ovesdotter-alm-2011-subjective,plank-2022-problem,Cabitza_Campagner_Basile_2023}. Extensive psychological research demonstrates the significant influence of both individual characteristics (e.g., age, gender, personality traits) and group-level factors (e.g., cultural background) on how we perceive and interpret emotions~\cite{,mesquita1992cultural, kring1998sex, costa2008revised, charles2010social}. Consequently, models trained on datasets with aggregated labels inevitably fail to capture the nuanced, individualized nature of affective responses. This limits their effectiveness in real-world applications that demand personalized understanding and responsiveness to diverse emotional expressions.

Limited attempts have been made to incorporate annotator background information~\cite{plaza-del-arco-etal-2024-emotion}.  For instance,~\citet{10.1145/3173574.3173986} provide demographic data alongside sentiment annotations in an online community dataset; however, this work is limited by its focus on sentiment (rather than fine-grained emotions), its restriction to a specific online community, and its lack of multi-faceted affective response measures. To our knowledge, no existing dataset combines comprehensive individual difference variables, fine-grained affective responses, and annotations of real-world news content, as ours does.

\section{Dataset Collection Protocol}
To address the limitations of existing emotion detection datasets and move towards more nuanced individual-level modeling, we develop a two-stage data collection protocol to capture individualized affective responses to news headlines (Figure~\ref{fig:data_collection}). Our protocol emphasizes ecological validity and the collection of rich persona variables to enable the study of how personal characteristics influence affective responses.

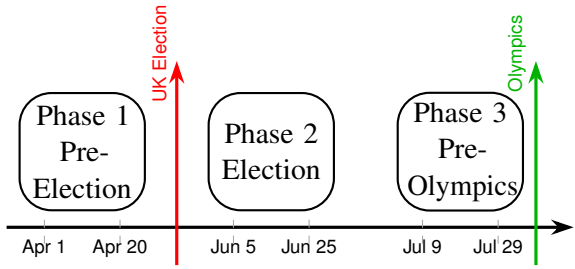
\begin{figure}[t!]
\centering
\begin{tikzpicture}[
    base/.style={draw, thick, black},
    phasebox/.style={base, fill=white, rounded corners=12pt, minimum height=1.55cm, align=center, text width=1.4cm}, %
    datelabel/.style={font=\scriptsize\sffamily, below=1pt}, %
    eventlabel/.style={font=\bfseries\sffamily, above=2pt}, %
    arrow/.style={-{Stealth[length=3mm, width=2mm]}, base, line width=1.2pt},
]

\draw[arrow] (0,0) -- (7.5,0);

\node[datelabel] (apr1) at (0.5,0) {Apr 1};
\node[datelabel] (apr20) at (1.5,0) {Apr 20};
\node[datelabel] (jun5) at (3,0) {Jun 5};
\node[datelabel] (jun25) at (4,0) {Jun 25};
\node[datelabel] (jul9) at (5.5,0) {Jul 9};
\node[datelabel] (jul29) at (6.5,0) {Jul 29};

\foreach \x in {0.5,1.5,3,4,5.5,6.5}
    \draw[dashed, gray!70] (\x,0.1) -- (\x,-0.3); %

\node[phasebox] (phase1) at (1,1) {Phase 1\\Pre-Election};
\node[phasebox] (phase2) at (3.5,1) {Phase 2\\Election};
\node[phasebox] (phase3) at (6,1) {Phase 3\\Pre-\\Olympics}; %

\draw[arrow, red] (2.25, -0.5) -- (2.25, 2.2); %
\node[rotate=90, above=0.1cm, font=\scriptsize\sffamily, red] at (2.25, 2.2) {UK Election};

\draw[arrow, green!70!black] (7, -0.5) -- (7, 2.2); %
\node[rotate=90, above=0.1cm, font=\scriptsize\sffamily, green!70!black] at (7, 2.2) {Olympics};

\end{tikzpicture}
\caption{Data collection timeline, with the 2024 UK General Election and 2024 Paris Olympics marked out.}
\label{fig:timeline}
\end{figure}

\paragraph{Sampling}
Our dataset comprises annotations of Facebook news posts, collected in three phases to capture diverse news contexts surrounding the 2024 UK general election and the Paris Olympics (see Figure~\ref{fig:timeline}). These phases are: Phase 1 (April 1-20), the pre-election period; Phase 2 (June 5-25), the election campaign period after Parliament's dissolution; and Phase 3 (July 9-29), the post-election and pre-Olympics period.

This three-phase design ensures temporal diversity and mitigates the influence of any single major event on our findings. We initially used random sampling (Phase 1) to gather a broad sample. Recognizing that some outlets are far more prolific but have lower engagement, we transitioned to stratified sampling (Phases 2 and 3) proportional to each outlet's follower count. This approach maximizes ecological validity by ensuring our sample reflects the news content that readers are actually likely to encounter.

For each phase, we collect news posts via CrowdTangle. While social media content may not represent an outlet's entire output, we posit that these posts reflect editorial choices and the outlet's intended public image. Each post typically includes an image, a short description, and the headline, with the image linking to the full article (see Figure~\ref{fig:sample} for an example). To ensure maximal ecological validity and minimize bias, we present screenshots instead of text of the posts to annotators, capturing reaction counts but excluding comments to avoid influencing annotator responses. The decision to use screenshots rather than just headlines is also supported by a pilot study (Section~\ref{sec:pre_study}) demonstrating significant differences in affective responses based on presentation modality.

\paragraph{Annotator Recruitment} We recruit annotators through Prolific, using quota sampling to ensure a relatively balanced representation across gender, age, political leaning, and UK geographical regions (see Figure~\ref{fig:geographic_representation} and Table~\ref{tab:persona_variable_breakdown} for details). Each of the 291 annotators contribute annotations for approximately 50 headlines. The annotation process takes around 45 minutes. Annotators are compensated £8.58, in accordance with the UK National Living Wage at the time of the data collection.

\paragraph{Stage 1: Persona Profile Survey}
This stage, implemented in Qualtrics, gathers background information (``persona variables'') about each annotator. The survey incorporates validated items from well-established questionnaires~\cite{ofcom2024newsconsumption,reuters2024dnr}, alongside standard psychological instruments. These variables (detailed in Table~\ref{tab:persona_variable_breakdown}) are selected to capture individual differences known to influence news interpretation and emotional responses, enabling us to study how these factors mediate affective reactions.

The collected variables span five key areas: \textbf{Demographics and Ideology} captures age, gender, and political ideology. \textbf{News Consumption and Trust} measures consumption patterns and trust ratings for major UK news outlets. \textbf{Cognitive Traits} are assessed through the Cognitive Reflection Test (CRT) \cite{frederick2005cognitive}. \textbf{Personality Traits} are measured using the 10-item Big Five Inventory (BFI-10) \cite{RAMMSTEDT2007203}. \textbf{Emotional Characteristics} are evaluated using both the Perth Emotional Reactivity Scale (PERS) \cite{preece2018assessing} and the Positive and Negative Affect Schedule (PANAS) \cite{crawford2004positive}.

\paragraph{Stage 2: Headline Annotation}
Annotators are provided with detailed guidelines (see Section \ref{sec:annotation_manual}), adapted from \citet{bradley2007affective}, which are accessible throughout the annotation process\footnote{Annotators are shown this link: \url{https://docs.google.com/document/d/1RPkjaPSksRbCy3y5d4WltidcUGhlH_np-aAuY2eH33c/}}. The annotation interface is built using the Potato annotation tool~\cite{pei-etal-2022-potato}.

We then present annotators with news posts public Facebook pages of major UK outlets (see Table~\ref{tab:summary_stats_per_outlet}). For each news post screenshot (see Figure~\ref{fig:screenshot} for exact question wording), annotators provide five types of responses: \textbf{Dimensional Emotion Ratings} capture valence, arousal, and dominance on a Likert scale of 1-7 using the Self-Assessment Manikin (SAM)~\cite{bradley1994measuring}. \textbf{Discrete Emotion Classifications} involve categorizing into one of Ekman's basic emotion~\cite{Ekman01051992}. \textbf{Modality Influence} assesses the relative influence of the image versus the text on their emotional response. \textbf{Personal Relevance} rates the headline's personal relevance. \textbf{Sharing Likelihood} measures the likelihood they would share the post. We randomize the order of the news posts.

\paragraph{Quality Control}
We recruit our annotators from Prolific, a platform recognized for having high participant attention and comprehension~\cite{Peer2022Data}. To further ensure data quality, we implement several procedural checks.

Annotators are required to spend at least 2 minutes on the instruction page (enforced by not showing the continuation button until 2 minutes’ time). The average time spent reviewing the instructions is 4.67 minutes, suggesting reasonable engagement beyond the minimum requirement.

We incorporate a comprehension check at the very beginning of the annotation task. Specifically, we present an excerpt from the annotation manual and ask two questions to see whether the annotators can actually understand our task. Those who fail to answer either one correctly are not allowed to proceed.

We include two attention check questions in our annotation job. Only 7 out of 291 annotators (2.4\%) fail either one of these checks, with several contacting us afterward acknowledging their errors. While these annotators are not excluded in the analysis, their IDs are flagged, allowing dataset users to filter them if desired.

To empirically validate that annotators understand and correctly apply the annotation guidelines and the annotation scales, we include three standardized Affective Norms for English Text (ANET) sentences as calibration items~\cite{bradley2007affective} in the annotation. Comparing our annotators' ratings on these items to the original ANET norms (converted to a 7-point scale) provides a direct measure of their ability to use the SAM scales consistently with the established protocol. Due to ANET licensing restrictions, the specific sentences cannot be disclosed. Table~\ref{tab:vad_comparison} presents a comparison of mean scores and standard deviations.

As shown in Table~\ref{tab:vad_comparison}, VAD scores from our UK-based annotator pool demonstrate strong correspondence with the original US-based ANET norms for these standardized stimuli. Mean differences are generally minor (typically <0.5 on the 7-point scale). This close alignment indicates that our annotators understand and apply the SAM scales in a manner consistent with established psychometric standards. While minor variations are expected due to demographic differences between our diverse UK sample and the original US university student sample, the overall consistency validates the integrity of our collected affective ratings.

\begin{table*}[t]
\centering
\small
\begin{tabular}{@{}l@{\hspace{1.5em}}cc@{\hspace{1.5em}}cc@{\hspace{1.5em}}cc@{}}
\toprule
\multirow{2}{*}{\textbf{Sentence ID}} & 
\multicolumn{2}{c}{\textbf{Valence}} & 
\multicolumn{2}{c}{\textbf{Arousal}} & 
\multicolumn{2}{c}{\textbf{Dominance}} \\
\cmidrule(lr){2-3} \cmidrule(lr){4-5} \cmidrule(l){6-7}
& \textbf{ANET} & \textbf{Ours} & \textbf{ANET} & \textbf{Ours} & \textbf{ANET} & \textbf{Ours} \\
\midrule
6400 & 2.17\,(1.24) & 1.55\,(0.88) & 6.32\,(1.12) & 6.19\,(1.05) & 2.91\,(1.91) & 2.57\,(1.45) \\
2550 & 6.08\,(0.98) & 6.30\,(0.84) & 2.15\,(1.54) & 1.81\,(1.24) & 5.10\,(1.39) & 4.38\,(0.95) \\
4450 & 6.46\,(0.83) & 6.17\,(1.02) & 5.91\,(1.36) & 5.58\,(1.14) & 4.50\,(1.58) & 4.91\,(1.22) \\
\bottomrule
\end{tabular}
\caption{Comparison of Valence (V), Arousal (A), and Dominance (D) ratings between ANET dataset and our annotations. Values show mean scores with standard deviations in parentheses.}
\label{tab:vad_comparison}

\end{table*}

\section{Descriptive Analysis}
\label{sec:descriptive_analysis}
Our dataset comprises 2,899 annotated news posts, with an average of 5.18 annotations per post from 291 distinct annotators.

\paragraph{Annotator Demographics}
Our annotator pool exhibits diversity across gender, political ideology, ethnicity, education levels, and other key demographic variables. Crucially, we have annotators from 97 out of 124 UK postcode areas, ensuring substantial geographic diversity within the UK. See Table~\ref{tab:persona_variable_breakdown} for a comprehensive breakdown of annotator characteristics.

\paragraph{Distribution of Annotations}
Figure~\ref{fig:all_dist} presents the distributions of the collected annotation variables. Key observations include: The neutral value (4) is the most frequent for all three dimensions. As expected, the valence scores tend to skew negatively, arousal scores are predominantly high, and dominance scores skew slightly low. For discrete emotions, ``neutral'' is the most commonly selected emotion, followed by ``sad''. Interestingly, the next most frequent emotion is "happy," which is likely due to the limitation of having only one category for positive emotions. Further details, including distributions for relevance, sharing likelihood, and modality influence, are available in Appendix Section~\ref{sec:auxiliary_variables_analysis}. Additionally, we analyze inter-annotator agreement in Section~\ref{sec:agreement}, finding Krippendorff's $\alpha$ values comparable to existing emotion annotation datasets, with moderate agreement for valence ($\alpha = 0.468$) and lower agreement for arousal ($\alpha = 0.145$) and dominance ($\alpha = 0.203$).

\paragraph{Outlet-level Analysis}

We present the summary statistics of affect annotations across news outlets in Table~\ref{tab:summary_stats_per_outlet}. All outlets are on average more negative content (low valence; with discrete emotions predominantly categorized as either neutral or sad/angry) while maintaining higher-than-neutral levels of arousal. See Section~\ref{sec:outlet_level_full} for a comparison between broadsheet and tabloid outlets.

\paragraph{News Post Characteristics}
To analyze the topical composition of news posts, we employ the IPTC NewsCode taxonomy \cite{iptc_media_topics}, a widely-adopted industry standard for news categorization. We choose this established taxonomy over topic modeling given the well-defined nature of news categorization as a task. We classify news post using zero-shot with Gemini 1.5 Pro (prompt in Section~\ref{sec:classification_prompt}). Figure~\ref{fig:topic_distribution} shows the topic distribution, and Figure~\ref{fig:mean_arousal_per_topic} shows mean arousal per topic. The most frequent categories are arts/culture/entertainment/media (25.4\%), crime/law/justice (12.9\%), and politics (9.6\%). This prevalence of hard news over soft news aligns with prior research on media organizations' social media strategies~\cite{Lamot21042022} and platform-specific characteristics of Facebook~\cite{newman2015reuters}. As expected, arousal is higher for topics like conflict/war (4.83) and disasters/accidents (4.77) compared to arts/culture (3.85), consistent with previous findings~\cite{soroka2019cross}.

\section{Regression Analysis}
\label{sec:regression_analysis}
To quantify the influence of individual differences on affective responses, and to assess the effectiveness of our collected persona variables in capturing these differences, we conduct a regression analysis using linear mixed-effects models, focusing on the arousal dimension as a case study (Likert scale, 1-7).

\paragraph{Models}
We construct three models to systematically decompose the variance in affective responses: (1) a \textbf{Null Model} with only news text as a random effect, serving as our baseline; (2) a \textbf{Persona Model} adding 47 persona variables as fixed effects while controlling for text effects; and (3) a \textbf{User Model} incorporating both news text and user ID as random effects to capture all user-level variance, including unobserved individual differences.

We evaluate each model using both marginal $R^2$ (variance explained by fixed effects) and conditional $R^2$ (variance explained by fixed and random effects) show the results in Table~\ref{tab:regression-results}.

\paragraph{Strong explanatory power of persona variables.} News content alone explains 13.1\% of the variance in arousal ratings (null model, conditional $R^2 = 0.131$). Incorporating our collected persona variables significantly increases the explained variance to 28.6\%. This improvement, higher than that observed in existing NLP datasets with annotator characteristics \cite{10.1145/3173574.3173986,hu-collier-2024-quantifying}, underscores the importance of individual differences in modeling subjective phenomena and validates the richness of the persona information collected in iNews.

\paragraph{Unobserved individual factors still matters.} Despite explicitly modeling a comprehensive set of persona variables, the User model explains more variance than the Persona model (0.317 vs. 0.286). This gap suggests the presence of additional unobserved individual factors that modulate affective responses—factors that remain unaccounted for even with our extensive variable collection.

\paragraph{Persona information matter in modeling affective responses.}
Our findings demonstrate that modeling individual differences is crucial for understanding affective responses to text. The persona variables collected in our iNews dataset capture a large portion of this individual variability, validating our data collection protocol and demonstrating the dataset's value for advancing personalized language technologies. The remaining unexplained variance highlights both the inherent complexity of human affect and the potential for future research to contextualize additional contributing factors.
\begin{table*}[t]
\small
\centering
\begin{tabular}{lcccc}
\toprule
\textbf{Model} & \textbf{Fixed Effects} & \textbf{Random Effects} & \textbf{Marginal $R^2$} & \textbf{Conditional $R^2$} \\
\midrule
Null & None  & Text & 0.000 & 0.131 \\
Persona & Persona variables (47) & Text & 0.152 & 0.286 \\
User & None  & Text + User & 0.000 & 0.317 \\
\bottomrule
\end{tabular}
\caption{Comparison of mixed-effects regression models for predicting affective arousal ratings. Marginal $R^2$ indicates variance explained by fixed effects alone, while Conditional $R^2$ shows the total variance explained by both fixed and random effects.}
\label{tab:regression-results}
\end{table*}

\section{Qualitative Analysis of Post-Annotation Questionnaire}
To complement our quantitative analysis of persona variables and gain a richer understanding of how individual differences shape emotional responses, we conduct a post-annotation qualitative study.  Twenty annotators from our main study complete an open-ended questionnaire (administered via Qualtrics/Prolific), consisting of six open-ended questions probing how readers process and respond to news content (see Section~\ref{sec:qual_interview} for questions and expanded analysis).

We perform a thematic analysis of the responses, employing a systematic coding approach facilitated by an LLM. The analysis reveals insights that help contextualize the individual differences observed in our survey data.  For instance, one annotator describe the influence of growing up during the cold war on their emotional responses. Another highlights how their working-class background leads them to be ``kind of numb to some types of news,'' while still emphasizing the emotional impact of ``people getting hurt for no reason.'' The news platform itself emerges as a mediating factor, with one participant stating, ``I don't really buy what I see on Facebook, so it doesn't get to me as much.`` These rich self-narratives, combined with structured persona data, could inform more nuanced models of individual differences in news response, aligning with recent work on using qualitative interview data to simulate human behavior \citep{park2024generativeagentsimulations1000}.

\section{Predicting Individual Affective Arousal}
Building on our regression analysis, we now investigate the capacity of current LLMs to predict individual-level affective response. We continue to focus on the emotional arousal dimension as a case study, examining how well models estimate specific annotators' responses under various zero-shot and few-shot conditions.

\subsection{Experimental Setup}
We randomly sample 30 annotators from iNews dataset. For each annotator, we reserve 32 of their annotated posts for potential few-shot demonstrations and utilize the remaining posts (579 samples total) for testing. For evaluation, we employ three complementary metrics that capture different aspects of prediction quality: Mean Absolute Error (MAE) to measure overall prediction accuracy, Exact Accuracy to identify precise matches with annotator ratings, and ±1 Accuracy (the percentage of predictions falling within one point of the ground truth) to account for the inherent subjectivity in emotional assessment by allowing slight variations. Our evaluation compares model predictions against each individual annotator's ratings.

We conduct experiments across seven frontier models, including both API-based models [Gemini 1.5 Pro~\cite{team2024gemini}, GPT-4o~\cite{hurst2024gpt}, Grok-2~\cite{xai2025grok2}] and open-weight models [Llama-3.2-90B-Vision~\cite{meta2024llama32}, Qwen2.5-VL-72B-Instruct~\cite{qwen2025vl}, Llama-3.3-70B-Instruct~\cite{llama2024llama33}, Llama-3.1-405B-Instruct~\cite{meta2024llama31}], with all except the last two capable of processing multimodal inputs. In rare cases where formatting or safety concerns prevents a model from generating a prediction, we assign -1 as the prediction to penalize such behavior. As we decode only a single token for the answer, temperature settings and sampling parameters are not relevant to this process.

We examine four input conditions. The text-only condition provides a detailed textual description of each news post, while the image-only condition uses the original news screenshot. We then augment each of these base conditions with persona information, creating text-with-persona and image-with-persona conditions where annotator characteristics are incorporated into the system prompt. Since our annotators originally rated news screenshots, we leverage Gemini 1.5 Pro to generate comprehensive textual descriptions for the text-only conditions, enriching these with headline text and engagement metrics. The complete prompt templates and an illustrative news post image-text pair are provided in Sections~\ref{sec:benchmark_prompt} and~\ref{sec:example_texual_description}, respectively. We present our zero-shot evaluation results in Table~\ref{tab:main_results}.

\begin{figure}[!h] %
    \centering
\includegraphics[width=0.9\linewidth]{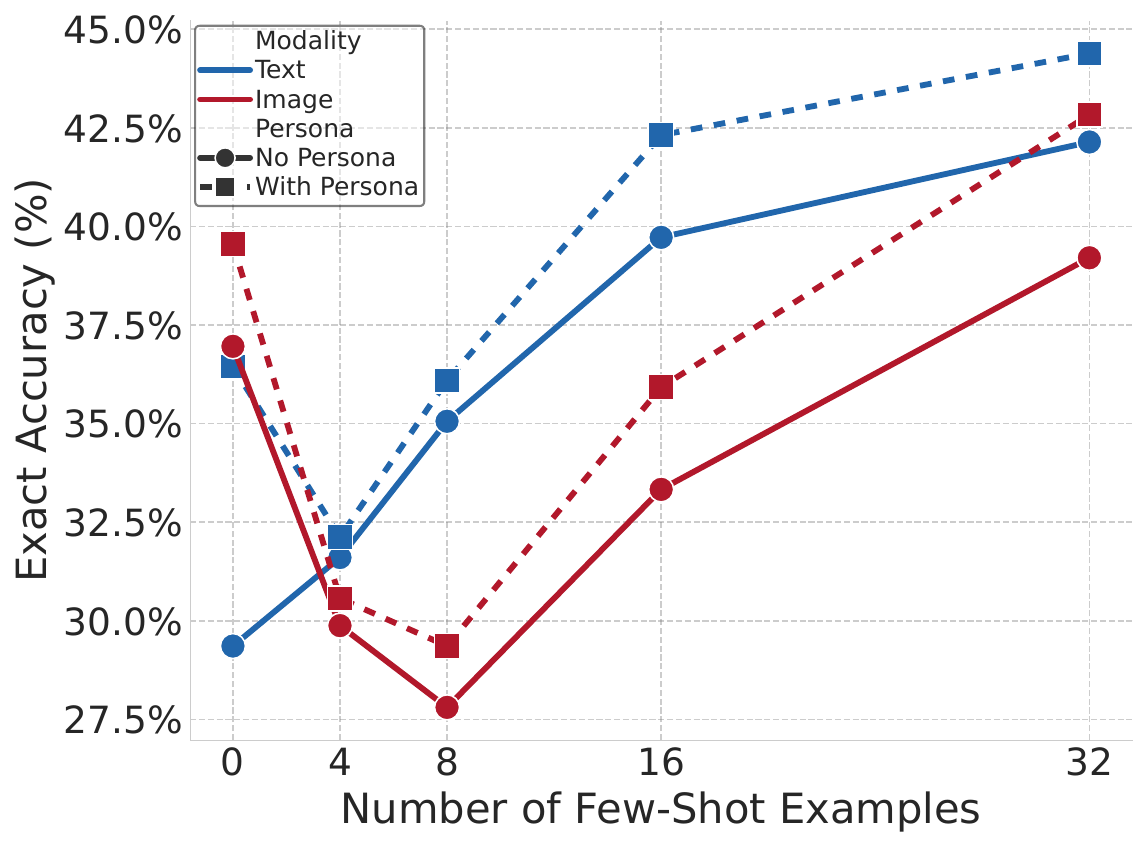}
\caption{Few-shot learning performance, measured by exact match accuracy (\%), as a function of the number of few-shot examples (0, 4, 8, 16, and 32).}
\label{fig:few_shot_main} %
\end{figure}

While persona variables provide valuable signals for personalization, they inevitably offer an incomplete view of individual preferences and behaviors, as the richness and complexity of human experience extends far beyond what can be captured through demographic and personality questionnaires~\cite{dong-etal-2024-llm}. We hypothesize that incorporating behavioral data, specifically, an individual's prior annotations, could provide complementary information for modeling affective responses. To test this hypothesis, we conduct $k$-shot experiments ($k \in \{4, 8, 16, 32\}$) with and without persona information across both text and image modalities. Figure~\ref{fig:few_shot_main} presents the Exact Accuracy results for Gemini 1.5 Pro, our top-performing zero-shot model (complete results in Table~\ref{tab:few_shot_full_results} and Figure~\ref{fig:all_subfigs_few_shot}). Due to resource constraints, we are only able to conduct few-shot experiments with Gemini 1.5 Pro.

\subsection{Zero-shot Evaluation}
\paragraph{Current LLMs demonstrate reasonable default zero-shot alignment with UK annotators.} Without personalization, models achieve seemingly encouraging baseline performance with ±1 accuracy exceeding 70\%. However, this metric alone can be misleading - a naive predictor that simply outputs the population mean would likely achieve similar ±1 accuracy given the roughly Gaussian distribution of arousal ratings (see Figure~\ref{fig:all_dist}). The consistently low exact accuracy (< 40\%) across all models provides a more stringent evaluation of true personalization capability. This substantial gap between ±1 and exact accuracy suggests that while models can broadly approximate the range of typical responses, they struggle to capture individual-specific variations in emotional reactions. 

\paragraph{Incorporating persona information consistently improves performance.} The improvements are particularly large for Gemini 1.5 Pro, where persona information reduces MAE by 11.6\% (1.034 → 0.914) for text input and 10.1\% (0.936 → 0.841) for image input. These substantial gains demonstrate that current LLMs can effectively leverage explicit persona variables to better simulate individual annotators, validating our persona variable collection strategy. This result aligns with prior work on the effectiveness of persona prompting~\cite{rescala-etal-2024-language, dong-etal-2024-llm, hu-collier-2024-quantifying}. 

\paragraph{Image inputs consistently outperform textual inputs.} Our analysis reveals a clear advantage for image-based prediction across all vision-language models except Llama-3.2-90B-Vision. The optimal performance is achieved by Gemini 1.5 Pro with image input and persona information (MAE: 0.841, ±1 Accuracy: 82.04\%), surpassing the best text-only configuration from Llama-3.1-405B-Instruct (MAE: 0.885, ±1 Accuracy: 81.17\%). This superiority of visual inputs aligns with prior working documenting stronger affective responses to images versus text in psychology and communication literature~\cite{https://doi.org/10.1002/ejsp.316,10.1111/jcom.12184}. Even our high-quality textual descriptions (example in Section~\ref{sec:example_texual_description}), appear unable to fully capture the affective richness encoded in visual stimuli. This observation is echoed by annotators who report particularly intense emotional reactions to images of suffering or tragedy (see Section~\ref{sec:open_ended_Q2}).

\paragraph{Models exhibit varying degrees of steerability through persona prompting.} Gemini 1.5 Pro, Grok-2,  Qwen2.5-VL-72B-Instruct and the Llama family show high responsiveness to persona information, while GPT-4o maintains more consistent behavior with and without persona information. This variation suggests fundamental differences in these models' capacity for steerably pluralistic alignment~\cite{sorensen2024position}.

\subsection{Few-shot Evaluation}
\paragraph{Few-shot learning demonstrates a consistent pattern of initial performance degradation followed by gradual recovery.} For both text and image modalities, we observe performance drop when transitioning from zero-shot to 4-shot setting. Performance gradually recovers with increasing demonstrations, with text-input models surpassing zero-shot performance at 8 shots (no persona) or 16 shots (with persona), continuing to improve up to 32 shots. However, the image modality shows a sharper initial decline and slower recovery, only matching zero-shot performance at 32 shots for exact accuracy and still lagging in ±1 Accuracy and MAE. This initial deterioration aligns with the early ascent (in terms of risk) phenomenon in in-context learning \cite{lin2024dual,agarwal2024manyshot}, where models initially struggle to effectively integrate limited demonstrations. We hypothesize that the inherent subjectivity and noise in emotional arousal annotations may exacerbate this effect, leading to overfitting with sparse examples before models learn to extract robust user-specific patterns.

\paragraph{Persona information provides consistent benefits across few-shot regimes.} Even at 32 shot, persona information yields substantial improvements (text: MAE 0.812 → 0.782, accuracy 0.421 → 0.444; image: MAE 0.926 → 0.858, accuracy 0.392 → 0.428). This persistent benefit suggests that explicit persona information captures complementary signals to those learned from demonstration examples. Drawing parallels to recommender systems literature, our few-shot approach is analogous to item-based recommendation, while persona prompting resembles natural-language-based recommendation [See~\citet{10.1145/3604915.3608845} for an overview]. Our results contribute to this line of research by demonstrating the potential value of hybrid approaches: while past behavior reveals specific preferences, persona information provides a broader context that may not be readily inferable from a limited set of behavioral examples.

\paragraph{Image few-shot prompting scales worse than text, despite zero-shot advantages.} While image inputs yield the best zero-shot performance, they exhibit both steeper initial performance degradation and more limited few-shot scaling compared to text inputs. Despite showing consistent improvements with additional demonstrations, image performance does not surpass the zero-shot level even at 32 shots. This pattern likely reflects both the increased complexity of visual processing and limitations of current vision-language models in few-shot learning scenarios. Although images contain the complete information available to human annotators, current VLMs appear unable to fully leverage this rich visual information in few-shot contexts, suggesting an area for future work.

\begin{table}[ht]
\small
\setlength{\tabcolsep}{4pt}
\centering
\begin{threeparttable}
\begin{tabular}{l@{\hspace{3pt}}lS[table-format=1.2]S[table-format=2.2]S[table-format=2.2]}
\toprule
& & \multicolumn{3}{c}{\textbf{Evaluation Metrics}} \\
\cmidrule(lr){3-5}
\textbf{Model} & \textbf{Input} & {\textbf{MAE}↓} & {\textbf{Acc}↑} & {\textbf{±1 Acc}↑} \\
\midrule
\multicolumn{5}{@{}l}{\textit{A. Vision-Language Models}} \\[0.5ex]
\multirow{4}{*}{Gemini-1.5 Pro} 
 & T & 1.03 & 29.36 & 74.44 \\
 & I & 0.94 & 36.96 & 77.03 \\ 
 & T+P & 0.91 & 36.44 & 78.76 \\ 
 & I+P & 0.84 & 39.55 & 82.04 \\ 
\cmidrule[0.4pt](l{0.25em}r{0.25em}){1-5}
\multirow{4}{*}{GPT-4o} 
 & T & 0.98 & 31.43 & 77.03 \\
 & I & 0.91 & 35.41 & 79.97 \\ 
 & T+P & 1.03 & 27.46 & 75.30 \\ 
 & I+P & 0.89 & 36.79 & 79.45 \\
\cmidrule[0.4pt](l{0.25em}r{0.25em}){1-5}
\multirow{4}{*}{Qwen2.5-VL-72B} 
 & T & 0.99 & 32.82 & 78.76 \\
 & I & 0.91 & 36.96 & 79.97\\
 & T+P & 0.92 & 34.72 & 80.31 \\
 & I+P & 0.90 & 37.48 & 80.48 \\
\cmidrule[0.4pt](l{0.25em}r{0.25em}){1-5}
\multirow{4}{*}{Llama-3.2-90B} 
 & T & 1.14 & 33.51 & 71.16 \\
 & I & 1.80 & 23.66 & 53.54 \\ 
 & T+P & 0.90 & 36.44 & 81.86\\ 
 & I+P & 1.31 & 22.28 & 64.25 \\ 
 \cmidrule[0.4pt](l{0.25em}r{0.25em}){1-5}
 \multirow{4}{*}{Grok-2}
 & T & 1.10 & 29.53 & 74.61 \\
 & I & 1.04 & 34.02 & 74.09 \\
 & T+P & 0.91 & 36.61 & 80.14 \\
 & I+P & 0.90 & 37.13 & 81.52 \\
\midrule
\multicolumn{5}{@{}l}{\textit{B. Language-Only Models}} \\[0.5ex]
\multirow{2}{*}{Llama-3.1-405B} 
 & T & 0.98 & 34.20 & 78.07 \\
 & T+P & 0.89 & 38.00 & 81.17 \\
\cmidrule[0.4pt](l{0.25em}r{0.25em}){1-5}
\multirow{2}{*}{Llama-3.3-70B} 
 & T & 1.16 & 31.61 & 71.16 \\
 & T+P & 0.94 & 37.31 & 79.62 \\
\bottomrule
\end{tabular}
\begin{tablenotes}[flushleft]
\small
\item \textbf{Input types:} T=Text, I=Image, P=Persona
\item \textbf{Metrics:} MAE=Mean Absolute Error (↓ lower is better), Acc=Exact Accuracy, ±1 Acc=Within-One Accuracy (↑ higher is better)
\end{tablenotes}

\end{threeparttable}
\caption{Zero-shot performance comparison across input modalities and models.}

\label{tab:main_results}

\end{table}

\section{Conclusion}
To facilitate the study of individual human behavior in response to stimuli, we introduce iNews, a novel dataset capturing individualized affective responses to news content, demonstrating that annotator characteristics could explain 15.2\% of response variance - higher than existing datasets. In our case study, we find that incorporating persona information consistently improves prediction accuracy (up to 7\% gains). We observe the early ascent phenomenon, where few-shot performance initially drops below zero-shot levels before recovering with additional examples. Notably, even at 32-shot (achieving 44.4\% accuracy), persona information continues to provide benefits, suggesting that explicit modeling of individual characteristics offers complementary signals to behavioral demonstrations. We also identify a modality gap: while image inputs excel in zero-shot settings, they show limited few-shot scaling compared to text inputs. Through its rich annotations and persona variables, iNews advances research in personalization, subjectivity, and affective computing, while providing new opportunities for studying human behavioral simulation.

\section{Limitations}
\paragraph{Annotator and Sample Scope} While iNews achieves substantial demographic and geographic diversity, with annotators from a majority of UK postcodes and various backgrounds, our sample most likely does not constitute a representative sample of the UK population or Facebook users. Nonetheless, iNews provides rich insights into how different demographic groups respond to news content. Our focus on UK-based annotators and UK news outlets necessarily constrains the applicability of our findings to other cultural, political, and media ecosystems. Future work should extend this work to diverse global contexts to build a more comprehensive understanding of news perception across different populations.

\paragraph{Methodological Considerations} Our emotion measurement framework, while grounded in established psychological constructs (VAD, Ekman), may not capture the full complexity of emotional responses to news content. Like most studies in this domain, we rely on self-reported emotions, which can be subject to various biases such as social desirability. Future work could strengthen the validity of these measurements by incorporating physiological measures (e.g., skin conductance, facial expressions) or triangulating multiple measurement approaches. Additionally, it is important to acknowledge that this paper is concerned with affective responses to news, which may differ in nature or intensity from emotions stemming from more direct, grounded lived experiences.

\paragraph{Platform Coverage.} We focused our data collection on Facebook posts, as Facebook remains the dominant social media news source in the UK as of 2024~\cite{ofcom2024newsconsumption_report}. While platform-specific effects may exist, our findings provide valuable insights into how users engage with news on a major distribution channel. Future work could extend this analysis to other platforms to understand platform-specific effects.

\paragraph{Data Quality} Although we implemented multiple quality control measures (attention checks, Captcha verification) and used Prolific's platform, which claims to provide 100\% genuine human participants\footnote{https://www.prolific.com/participant-pool}, we cannot completely rule out the possibility of AI-generated responses. Our modeling results support the high quality of the collected annotations, though as with any large-scale annotation effort, maintaining perfect attention throughout all responses cannot be guaranteed.

\section{Ethical Considerations}
This study was conducted with the approval of our institutional ethics review board. All annotators provided informed consent before participation and were compensated fairly according to UK National Living Wage. No personally identifiable information was collected in our dataset. To protect participant privacy, we paraphrase open-ended responses quoted in this paper while preserving their essential meaning. During data collection, Prolific IDs were temporarily used to link annotations with persona data, but the publicly released dataset will contain only newly generated, anonymized participant IDs. Given our focus on UK-based annotators and news sources, we recognize the inherent limitations in global generalizability. However, we made conscious efforts to ensure demographic diversity within our annotator pool through Prolific's stratified sampling features. We acknowledge that emotional responses to news can be culturally specific and have thoroughly documented our annotator demographics to enable future researchers to account for potential demographic skews in their analyses. 

\section*{Acknowledgments}
This work was funded by the Cambridge Language Sciences Incubator Fund.
T.H is supported by Gates Cambridge Trust (grant OPP1144 from the Bill \& Melinda Gates Foundation). We  acknowledge the compute support from Google DeepMind. We thank Kiran Garimella, Wen Wu, Hannah Claus, Songbo Hu, Chengzu Li, Joe Watson, Yara Kyrychenko, Paul Röttger, Jon Roozenbeek, Hope Schroeder, River Yijiang Dong, Flor Miriam Plaza-del-Arco, Donya Rooein, Debora Nozza, Arianna Muti, Yinhong Liu, Hannah Sterz and David Stillwell, the EPI-AI team, and the ACL 2024 student research workshop for helpful feedback and discussions at various stages of the project.

\bibliography{custom}

\newpage
\appendix
\nopagebreak
\onecolumn
\section{Appendix}
\subsection{Full Persona Variables}
\footnotesize{
\begin{longtable}{p{6.2cm}rl}
\caption{\small \textbf{Full Persona Variables Breakdowns.} 
Counts and percentages of participants by persona variables. Big-Five Personality traits are assessed using the BFI-10 scale~\cite{RAMMSTEDT2007203}. For the Perth Emotional Reactivity Scale~\cite{preece2018assessing}, we include one question each from four dimensions (negative-activation, negative-intensity, positive-activation, positive-intensity) due to questionnaire length constraint~\cite{preece2018assessing}. Additionally, we include net positive scores calculated from ten items on the Positive and Negative Affect Schedule~\cite{crawford2004positive}. Our dataset includes a wide range of persona variables at both group and individual levels.}
\label{tab:persona_variable_breakdown} \\
\endfirsthead
\caption[]{\small \textbf{Full Persona Variables Breakdowns.} Counts and percentages of participants by persona variables. Big-Five Personality traits are assessed using the BFI-10 scale~\cite{RAMMSTEDT2007203}. For the Perth Emotional Reactivity Scale~\cite{preece2018assessing}, we include one question each from four dimensions (negative-activation, negative-intensity, positive-activation, positive-intensity) due to questionnaire length constraint~\cite{preece2018assessing}. Additionally, we include net positive scores calculated from ten items on the Positive and Negative Affect Schedule~\cite{crawford2004positive}. Our dataset includes a wide range of persona variables at both group and individual levels.} \\
\toprule
\midrule
\endhead
\midrule
\multicolumn{3}{r}{Continued on next page} \\
\midrule
\endfoot
\bottomrule
\endlastfoot
\midrule
{\cellcolor{white}}\textbf{Age} & {\cellcolor{white}}  & {\cellcolor{white}}  \\
\midrule
{\cellcolor[HTML]{F2F2F2}}  18-24 years old & {\cellcolor[HTML]{F2F2F2}} 23 & {\cellcolor[HTML]{F2F2F2}} \pcb{7.9}\% \\

{\cellcolor{white}}  25-34 years old & {\cellcolor{white}} 89 & {\cellcolor{white}} \pcb{30.6}\% \\
{\cellcolor{white}}  35-44 years old & {\cellcolor{white}} 77 & {\cellcolor{white}} \pcb{26.5}\% \\
{\cellcolor[HTML]{F2F2F2}} 45-54 years old & {\cellcolor[HTML]{F2F2F2}} 49 & {\cellcolor[HTML]{F2F2F2}} \pcb{16.8}\% \\
{\cellcolor{white}} 55-64 years old & {\cellcolor{white}} 37 & {\cellcolor{white}} \pcb{12.7}\% \\
{\cellcolor[HTML]{F2F2F2}} 65+ years old & {\cellcolor[HTML]{F2F2F2}} 16 & {\cellcolor[HTML]{F2F2F2}} \pcb{5.5}\% \\
\midrule
{\cellcolor{white}}\textbf{Sex} & {\cellcolor{white}}  & {\cellcolor{white}}  \\
\midrule
{\cellcolor{white}} Male & {\cellcolor{white}} 152 & {\cellcolor{white}} \pcb{52.2}\% \\
{\cellcolor[HTML]{F2F2F2}} Female & {\cellcolor[HTML]{F2F2F2}} 139 & {\cellcolor[HTML]{F2F2F2}} \pcb{47.8}\% \\
\midrule
\textbf{LGBTQ+ Self-Identification} & {\cellcolor{white}}  & {\cellcolor{white}}  \\
\midrule
{\cellcolor{white}} Yes & {\cellcolor{white}} 253 & {\cellcolor{white}} \pcb{86.9}\% \\
{\cellcolor[HTML]{F2F2F2}} No & {\cellcolor[HTML]{F2F2F2}} 38 & {\cellcolor[HTML]{F2F2F2}} \pcb{13.1}\% \\
\midrule

{\cellcolor{white}}\textbf{Ethnicity (Simplified)
} & {\cellcolor{white}}  & {\cellcolor{white}}  \\
\midrule
{\cellcolor{white}} White & {\cellcolor{white}} 239 & {\cellcolor{white}} \pcb{82.1}\% \\
{\cellcolor[HTML]{F2F2F2}} Mixed & {\cellcolor[HTML]{F2F2F2}} 20 & {\cellcolor[HTML]{F2F2F2}} \pcb{6.9}\% \\
{\cellcolor{white}} Asian & {\cellcolor{white}} 17 & {\cellcolor{white}} \pcb{5.8}\% \\
{\cellcolor[HTML]{F2F2F2}} Black & {\cellcolor[HTML]{F2F2F2}} 15 & {\cellcolor[HTML]{F2F2F2}} \pcb{5.2}\% \\

\midrule
{\cellcolor{white}}\textbf{Personal Income (GBP)}
 & {\cellcolor{white}}  & {\cellcolor{white}}  \\
\midrule
{\cellcolor{white}} Less than £10,000 & {\cellcolor{white}} 57 & {\cellcolor{white}} \pcb{19.6}\% \\
{\cellcolor[HTML]{F2F2F2}} £10,000 - £19,999 & {\cellcolor[HTML]{F2F2F2}} 60 & {\cellcolor[HTML]{F2F2F2}} \pcb{20.6}\% \\
{\cellcolor{white}} £20,000 - £29,999 & {\cellcolor{white}} 72 & {\cellcolor{white}} \pcb{24.7}\% \\
{\cellcolor[HTML]{F2F2F2}} £30,000 - £39,999 & {\cellcolor[HTML]{F2F2F2}} 49 & {\cellcolor[HTML]{F2F2F2}} \pcb{16.8}\% \\
{\cellcolor{white}} £40,000 - £49,999 & {\cellcolor{white}} 24 & {\cellcolor{white}} \pcb{8.25}\% \\
{\cellcolor[HTML]{F2F2F2}} £50,000 - £59,999 & {\cellcolor[HTML]{F2F2F2}} 12 & {\cellcolor[HTML]{F2F2F2}} \pcb{4.12}\% \\
{\cellcolor{white}} £60,000 - £69,999 & {\cellcolor{white}} 6 & {\cellcolor{white}} \pcb{2.06}\% \\
{\cellcolor[HTML]{F2F2F2}} £70,000 - £79,999 & {\cellcolor[HTML]{F2F2F2}} 3 & {\cellcolor[HTML]{F2F2F2}} \pcb{1.03}\% \\
{\cellcolor{white}} £80,000 - £89,999 & {\cellcolor{white}} 5 & {\cellcolor{white}} \pcb{1.72}\% \\
{\cellcolor[HTML]{F2F2F2}} £90,000 - £99,999 & {\cellcolor[HTML]{F2F2F2}} 2 & {\cellcolor[HTML]{F2F2F2}} \pcb{0.687}\% \\
{\cellcolor{white}} More than £150,000 & {\cellcolor{white}} 1 & {\cellcolor{white}} \pcb{0.344}\% \\
\midrule

{\cellcolor{white}}\textbf{Highest Education Level Completed} & {\cellcolor{white}}  & {\cellcolor{white}}  \\
\midrule
{\cellcolor{white}} No formal qualifications & {\cellcolor{white}} 3 & {\cellcolor{white}} \pcb{1.03}\% \\
{\cellcolor[HTML]{F2F2F2}} Secondary education (e.g. GED/GCSE) & {\cellcolor[HTML]{F2F2F2}} 24 & {\cellcolor[HTML]{F2F2F2}} \pcb{8.25}\% \\
{\cellcolor{white}} High school diploma/A-levels & {\cellcolor{white}} 49 & {\cellcolor{white}} \pcb{16.8}\% \\
{\cellcolor[HTML]{F2F2F2}} Technical/community college & {\cellcolor[HTML]{F2F2F2}} 38 & {\cellcolor[HTML]{F2F2F2}} \pcb{13.1}\% \\
{\cellcolor{white}} Undergraduate degree (BA/BSc/other) & {\cellcolor{white}} 124 & {\cellcolor{white}} \pcb{42.6}\% \\
{\cellcolor[HTML]{F2F2F2}} Graduate degree (MA/MSc/MPhil/other) & {\cellcolor[HTML]{F2F2F2}} 49 & {\cellcolor[HTML]{F2F2F2}} \pcb{16.8}\% \\
{\cellcolor{white}} Doctorate degree (PhD/other) & {\cellcolor{white}} 4 & {\cellcolor{white}} \pcb{1.37}\% \\
\midrule
\textbf{Are you a student?
} & {\cellcolor{white}}  & {\cellcolor{white}}  \\
\midrule
{\cellcolor{white}} Yes & {\cellcolor{white}} 34 & {\cellcolor{white}} \pcb{11.7}\% \\
{\cellcolor[HTML]{F2F2F2}} No & {\cellcolor[HTML]{F2F2F2}} 246 & {\cellcolor[HTML]{F2F2F2}} \pcb{84.5}\% \\
{\cellcolor{white}} DATA EXPIRED & {\cellcolor{white}} 11 & {\cellcolor{white}} \pcb{3.8}\% \\

\midrule
{\cellcolor{white}}\textbf{Employment Status} & {\cellcolor{white}} & {\cellcolor{white}} \\
\midrule
{\cellcolor{white}} Due to start a new job within the next month & {\cellcolor{white}} 3 & {\cellcolor{white}} \pcb{1.03}\% \\
{\cellcolor[HTML]{F2F2F2}} Other & {\cellcolor[HTML]{F2F2F2}} 6 & {\cellcolor[HTML]{F2F2F2}} \pcb{2.06}\% \\
{\cellcolor{white}} DATA EXPIRED & {\cellcolor{white}} 12 & {\cellcolor{white}} \pcb{4.12}\% \\
{\cellcolor[HTML]{F2F2F2}} Unemployed (and job seeking) & {\cellcolor[HTML]{F2F2F2}} 13 & {\cellcolor[HTML]{F2F2F2}} \pcb{4.47}\% \\
{\cellcolor{white}} Not in paid work (e.g. homemaker, retired or disabled) & {\cellcolor{white}} 36 & {\cellcolor{white}} \pcb{12.4}\% \\
{\cellcolor[HTML]{F2F2F2}} Part-Time & {\cellcolor[HTML]{F2F2F2}} 50 & {\cellcolor[HTML]{F2F2F2}} \pcb{17.2}\% \\
{\cellcolor{white}} Full-Time & {\cellcolor{white}} 171 & {\cellcolor{white}} \pcb{58.8}\% \\
\midrule
{\cellcolor{white}}\textbf{Nationality (UK)} & {\cellcolor{white}} & {\cellcolor{white}} \\
\midrule
{\cellcolor{white}} England & {\cellcolor{white}} 232 & {\cellcolor{white}} \pcb{79.7}\% \\
{\cellcolor[HTML]{F2F2F2}} Scotland & {\cellcolor[HTML]{F2F2F2}} 30 & {\cellcolor[HTML]{F2F2F2}} \pcb{10.3}\% \\
{\cellcolor{white}} Wales & {\cellcolor{white}} 18 & {\cellcolor{white}} \pcb{6.19}\% \\
{\cellcolor[HTML]{F2F2F2}} Northern Ireland & {\cellcolor[HTML]{F2F2F2}} 11 & {\cellcolor[HTML]{F2F2F2}} \pcb{3.78}\% \\
\midrule
{\cellcolor{white}}\textbf{Political Leaning
} & {\cellcolor{white}}  & {\cellcolor{white}}  \\
\midrule
{\cellcolor{white}} Centre & {\cellcolor{white}} 118 & {\cellcolor{white}} \pcb{40.5}\% \\
{\cellcolor[HTML]{F2F2F2}} Right & {\cellcolor[HTML]{F2F2F2}} 84 & {\cellcolor[HTML]{F2F2F2}} \pcb{28.9}\% \\
{\cellcolor{white}} Left & {\cellcolor{white}} 82 & {\cellcolor{white}} \pcb{28.2}\% \\
{\cellcolor[HTML]{F2F2F2}} DATA EXPIRED & {\cellcolor[HTML]{F2F2F2}} 7 & {\cellcolor[HTML]{F2F2F2}} \pcb{2.4}\% \\
\midrule
{\cellcolor{white}}\textbf{How interested, if at all, would you say you are in politics and the news?} & {\cellcolor{white}}  & {\cellcolor{white}}  \\
\midrule
{\cellcolor{white}} Not at all interested & {\cellcolor{white}} 3 & {\cellcolor{white}} \pcb{1.03}\% \\
{\cellcolor[HTML]{F2F2F2}} Not very interested & {\cellcolor[HTML]{F2F2F2}} 30 & {\cellcolor[HTML]{F2F2F2}} \pcb{10.3}\% \\
{\cellcolor{white}} Somewhat interested & {\cellcolor{white}} 113 & {\cellcolor{white}} \pcb{38.8}\% \\
{\cellcolor[HTML]{F2F2F2}} Very interested & {\cellcolor[HTML]{F2F2F2}} 102 & {\cellcolor[HTML]{F2F2F2}} \pcb{35.1}\% \\
{\cellcolor{white}} Extremely interested & {\cellcolor{white}} 43 & {\cellcolor{white}} \pcb{14.8}\% \\
\midrule
{\cellcolor{white}}\textbf{In the past week, on average, how much time per day did you spend consuming news from all sources (online, TV, print, radio, etc.)?} & {\cellcolor{white}}  & {\cellcolor{white}}  \\
\midrule
{\cellcolor{white}} Less than 15 minutes & {\cellcolor{white}} 14 & {\cellcolor{white}} \pcb{4.81}\% \\
{\cellcolor[HTML]{F2F2F2}} 15 minutes to less than 30 minutes & {\cellcolor[HTML]{F2F2F2}} 45 & {\cellcolor[HTML]{F2F2F2}} \pcb{15.5}\% \\
{\cellcolor{white}} 30 minutes to less than 1 hour & {\cellcolor{white}} 94 & {\cellcolor{white}} \pcb{32.3}\% \\
{\cellcolor[HTML]{F2F2F2}} 1 hour to less than 2 hours & {\cellcolor[HTML]{F2F2F2}} 77 & {\cellcolor[HTML]{F2F2F2}} \pcb{26.5}\% \\
{\cellcolor{white}} 2 hours or more & {\cellcolor{white}} 61 & {\cellcolor{white}} \pcb{21.0}\% \\
\midrule
{\cellcolor{white}}\textbf{How confident are you in your ability to distinguish between reliable and unreliable news sources?} & {\cellcolor{white}}  & {\cellcolor{white}}  \\
\midrule
{\cellcolor{white}} Not at all confident & {\cellcolor{white}} 3 & {\cellcolor{white}} \pcb{1.03}\% \\
{\cellcolor[HTML]{F2F2F2}} Slightly confident & {\cellcolor[HTML]{F2F2F2}} 37 & {\cellcolor[HTML]{F2F2F2}} \pcb{12.7}\% \\
{\cellcolor{white}} Moderately confident & {\cellcolor{white}} 109 & {\cellcolor{white}} \pcb{37.5}\% \\
{\cellcolor[HTML]{F2F2F2}} Quite confident & {\cellcolor[HTML]{F2F2F2}} 123 & {\cellcolor[HTML]{F2F2F2}} \pcb{42.3}\% \\
{\cellcolor{white}} Completely confident & {\cellcolor{white}} 19 & {\cellcolor{white}} \pcb{6.5}\% \\
\midrule
{\cellcolor{white}}\textbf{How often do you fact-check news stories you come across?} & {\cellcolor{white}}  & {\cellcolor{white}}  \\
\midrule
{\cellcolor{white}} Never & {\cellcolor{white}} 14 & {\cellcolor{white}} \pcb{4.81}\% \\
{\cellcolor[HTML]{F2F2F2}} Rarely & {\cellcolor[HTML]{F2F2F2}} 55 & {\cellcolor[HTML]{F2F2F2}} \pcb{18.9}\% \\
{\cellcolor{white}} Often & {\cellcolor{white}} 84 & {\cellcolor{white}} \pcb{28.9}\% \\
{\cellcolor[HTML]{F2F2F2}} Sometimes & {\cellcolor[HTML]{F2F2F2}} 132 & {\cellcolor[HTML]{F2F2F2}} \pcb{45.4}\% \\
{\cellcolor{white}} Always & {\cellcolor{white}} 6 & {\cellcolor{white}} \pcb{2.06}\% \\
\midrule
{\cellcolor{white}}\textbf{When reading a news article, how often do you consider the author's potential biases or agenda?} & {\cellcolor{white}}  & {\cellcolor{white}}  \\
\midrule
{\cellcolor{white}} Never & {\cellcolor{white}} 3 & {\cellcolor{white}} \pcb{1.03}\% \\
{\cellcolor[HTML]{F2F2F2}} Rarely & {\cellcolor[HTML]{F2F2F2}} 27 & {\cellcolor[HTML]{F2F2F2}} \pcb{9.28}\% \\
{\cellcolor{white}} Often & {\cellcolor{white}} 112 & {\cellcolor{white}} \pcb{38.5}\% \\
{\cellcolor[HTML]{F2F2F2}} Sometimes & {\cellcolor[HTML]{F2F2F2}} 105 & {\cellcolor[HTML]{F2F2F2}} \pcb{36.1}\% \\
{\cellcolor{white}} Always & {\cellcolor{white}} 44 & {\cellcolor{white}} \pcb{15.1}\% \\
\midrule

{\cellcolor{white}}\textbf{To what extent do you agree or disagree with the following statement: I often get emotionally affected by the news I read.} & {\cellcolor{white}}  & {\cellcolor{white}}  \\
\midrule
 Strongly disagree & 6 & \pcb{2.06}\% \\
\rowcolor[HTML]{F2F2F2} Disagree & 47 & \pcb{16.2}\% \\
 Neither agree nor disagree & 62 & \pcb{21.3}\% \\
\rowcolor[HTML]{F2F2F2} Agree & 157 & \pcb{54.0}\% \\
 Strongly agree & 19 & \pcb{6.53}\% \\
\midrule

{\cellcolor{white}}\textbf{PLEASE SELECT ALL THAT APPLY:
Which of the following platforms do you use for news nowadays?} & {\cellcolor{white}}  & {\cellcolor{white}}  \\
\midrule
\cellcolor{white} Radio & \cellcolor{white} 270 & \cellcolor{white} \pcb{92.8}\% \\
\cellcolor[HTML]{F2F2F2} Website & \cellcolor[HTML]{F2F2F2} 263 & \cellcolor[HTML]{F2F2F2} \pcb{90.4}\% \\
\cellcolor{white} Social media & \cellcolor{white} 217 & \cellcolor{white} \pcb{74.6}\% \\
\cellcolor[HTML]{F2F2F2} Television & \cellcolor[HTML]{F2F2F2} 203 & \cellcolor[HTML]{F2F2F2} \pcb{69.8}\% \\
\cellcolor{white} Word of mouth & \cellcolor{white} 164 & \cellcolor{white} \pcb{56.4}\% \\
\cellcolor[HTML]{F2F2F2} Podcasts & \cellcolor[HTML]{F2F2F2} 88 & \cellcolor[HTML]{F2F2F2} \pcb{30.2}\% \\
\cellcolor{white} Print newspaper / magazines & \cellcolor{white} 75 & \cellcolor{white} \pcb{25.8}\% \\
\cellcolor[HTML]{F2F2F2} I don't & \cellcolor[HTML]{F2F2F2} 0 & \cellcolor[HTML]{F2F2F2} \pcb{0}\% \\
\midrule
{\cellcolor{white}}\textbf{Which of the following social media sites do you use on a regular basis (at least once a month)? Choose any that apply.} & {\cellcolor{white}}  & {\cellcolor{white}}  \\
\midrule
{\cellcolor{white}} Youtube & {\cellcolor{white}} 243 & {\cellcolor{white}} \pcb{83.5}\% \\
{\cellcolor[HTML]{F2F2F2}} Facebook & {\cellcolor[HTML]{F2F2F2}} 218 & {\cellcolor[HTML]{F2F2F2}} \pcb{74.9}\% \\
{\cellcolor{white}} Instagram & {\cellcolor{white}} 171 & {\cellcolor{white}} \pcb{58.8}\% \\
{\cellcolor[HTML]{F2F2F2}} Twitter & {\cellcolor[HTML]{F2F2F2}} 150 & {\cellcolor[HTML]{F2F2F2}} \pcb{51.5}\% \\
{\cellcolor{white}} Reddit & {\cellcolor{white}} 119 & {\cellcolor{white}} \pcb{40.9}\% \\
{\cellcolor[HTML]{F2F2F2}} Linkedin & {\cellcolor[HTML]{F2F2F2}} 97 & {\cellcolor[HTML]{F2F2F2}} \pcb{33.3}\% \\
{\cellcolor{white}} TikTok & {\cellcolor{white}} 69 & {\cellcolor{white}} \pcb{23.7}\% \\
{\cellcolor[HTML]{F2F2F2}} Pinterest & {\cellcolor[HTML]{F2F2F2}} 62 & {\cellcolor[HTML]{F2F2F2}} \pcb{21.3}\% \\
{\cellcolor{white}} Snapchat & {\cellcolor{white}} 48 & {\cellcolor{white}} \pcb{16.5}\% \\
{\cellcolor[HTML]{F2F2F2}} Google Plus & {\cellcolor[HTML]{F2F2F2}} 15 & {\cellcolor[HTML]{F2F2F2}} \pcb{5.15}\% \\
{\cellcolor{white}} Tumblr & {\cellcolor{white}} 13 & {\cellcolor{white}} \pcb{4.47}\% \\
{\cellcolor[HTML]{F2F2F2}} Meetup & {\cellcolor[HTML]{F2F2F2}} 9 & {\cellcolor[HTML]{F2F2F2}} \pcb{3.09}\% \\
{\cellcolor{white}} Flickr & {\cellcolor{white}} 5 & {\cellcolor{white}} \pcb{1.72}\% \\
{\cellcolor[HTML]{F2F2F2}} Medium & {\cellcolor[HTML]{F2F2F2}} 3 & {\cellcolor[HTML]{F2F2F2}} \pcb{1.03}\% \\
{\cellcolor{white}} VK & {\cellcolor{white}} 3 & {\cellcolor{white}} \pcb{1.03}\% \\
{\cellcolor[HTML]{F2F2F2}} Vine.co & {\cellcolor[HTML]{F2F2F2}} 2 & {\cellcolor[HTML]{F2F2F2}} \pcb{0.687}\% \\
\midrule
{\cellcolor{white}}\textbf{How trustworthy or untrustworthy do you rate the news reported by the following media organizations? - BBC} & {\cellcolor{white}}  & {\cellcolor{white}}  \\
\midrule
 Don't know & 0 & \pcb{0}\% \\
\rowcolor[HTML]{F2F2F2} Very untrustworthy & 43 & \pcb{14.4}\% \\
Untrustworthy & 37 & \pcb{12.3}\% \\
\rowcolor[HTML]{F2F2F2} Neither trustworthy nor untrustworthy & 28 & \pcb{9.3}\% \\
 Trustworthy & 140 & \pcb{46.7}\% \\
\rowcolor[HTML]{F2F2F2}  Very trustworthy & 52 & \pcb{17.3}\% \\
\midrule
{\cellcolor{white}}\textbf{How trustworthy or untrustworthy do you rate the news reported by the following media organizations? - The Financial Times} & {\cellcolor{white}}  & {\cellcolor{white}}  \\
\midrule
Don't know & 20 & \pcb{6.87}\% \\
\rowcolor[HTML]{F2F2F2}Very untrustworthy & 11 & \pcb{3.78}\% \\
Untrustworthy & 17 & \pcb{5.84}\% \\
\rowcolor[HTML]{F2F2F2} Neither trustworthy nor untrustworthy & 49 & \pcb{16.8}\% \\
Trustworthy & 156 & \pcb{53.6}\% \\
\rowcolor[HTML]{F2F2F2} Very trustworthy & 38 & \pcb{13.1}\% \\
\midrule
{\cellcolor{white}}\textbf{How trustworthy or untrustworthy do you rate the news reported by the following media organizations? - The Guardian} & {\cellcolor{white}}  & {\cellcolor{white}}  \\
\midrule
 Don't know & 13 & \pcb{4.47}\% \\
\rowcolor[HTML]{F2F2F2} Very untrustworthy & 24 & \pcb{8.25}\% \\
 Untrustworthy & 23 & \pcb{7.90}\% \\
\rowcolor[HTML]{F2F2F2} Neither trustworthy nor 
 untrustworthy & 55 & \pcb{18.9}\% \\
 Trustworthy & 152 & \pcb{52.2}\% \\
\rowcolor[HTML]{F2F2F2}  Very trustworthy & 24 & \pcb{8.25}\% \\
\midrule
{\cellcolor{white}}\textbf{How trustworthy or untrustworthy do you rate the news reported by the following media organizations? - The Times \& Sunday Times} & {\cellcolor{white}}  & {\cellcolor{white}}  \\
\midrule
 Don't know & 19 & \pcb{6.53}\% \\
\rowcolor[HTML]{F2F2F2}  Very untrustworthy & 17 & \pcb{5.84}\% \\
Untrustworthy & 33 & \pcb{11.3}\% \\
\rowcolor[HTML]{F2F2F2}Neither trustworthy nor untrustworthy & 75 & \pcb{25.8}\% \\
Trustworthy & 130 & \pcb{44.7}\% \\
\rowcolor[HTML]{F2F2F2}Very trustworthy & 17 & \pcb{5.84}\% \\
\midrule
{\cellcolor{white}}\textbf{How trustworthy or untrustworthy do you rate the news reported by the following media organizations? - The Independent} & {\cellcolor{white}}  & {\cellcolor{white}}  \\
\midrule
Don't know & 15 & \pcb{5.15}\% \\
\rowcolor[HTML]{F2F2F2}Very untrustworthy & 13 & \pcb{4.47}\% \\
Untrustworthy & 32 & \pcb{11.0}\% \\
\rowcolor[HTML]{F2F2F2}Neither trustworthy nor untrustworthy & 79 & \pcb{27.1}\% \\
Trustworthy & 137 & \pcb{47.1}\% \\
\rowcolor[HTML]{F2F2F2} Very trustworthy & 15 & \pcb{5.15}\% \\
\midrule

{\cellcolor{white}}\textbf{How trustworthy or untrustworthy do you rate the news reported by the following media organizations? - Sky} & {\cellcolor{white}}  & {\cellcolor{white}}  \\
\midrule
\rowcolor[HTML]{F2F2F2} Don't know & 7 & \pcb{2.41}\% \\
Very untrustworthy & 36 & \pcb{12.4}\% \\
\rowcolor[HTML]{F2F2F2}Untrustworthy & 59 & \pcb{20.3}\% \\
Neither trustworthy nor untrustworthy & 76 & \pcb{26.1}\% \\
\rowcolor[HTML]{F2F2F2}Trustworthy & 98 & \pcb{33.7}\% \\
Very trustworthy & 15 & \pcb{5.15}\% \\
\midrule

{\cellcolor{white}}\textbf{How trustworthy or untrustworthy do you rate the news reported by the following media organizations? - The Economist} & {\cellcolor{white}}  & {\cellcolor{white}}  \\
\midrule
Don't know & 29 & \pcb{9.97}\% \\
\rowcolor[HTML]{F2F2F2}Very untrustworthy & 10 & \pcb{3.44}\% \\
Untrustworthy & 24 & \pcb{8.25}\% \\
\rowcolor[HTML]{F2F2F2}Neither trustworthy nor untrustworthy & 69 & \pcb{23.7}\% \\
Trustworthy & 135 & \pcb{46.4}\% \\
\rowcolor[HTML]{F2F2F2} Very trustworthy & 24 & \pcb{8.25}\% \\
\midrule

{\cellcolor{white}}\textbf{How trustworthy or untrustworthy do you rate the news reported by the following media organizations? - The Telegraph} & {\cellcolor{white}} & {\cellcolor{white}} \\
\midrule
\rowcolor[HTML]{F2F2F2} Don't know & 13 & \pcb{4.47}\% \\
Very untrustworthy & 24 & \pcb{8.25}\% \\
\rowcolor[HTML]{F2F2F2} Untrustworthy & 47 & \pcb{16.2}\% \\
Neither trustworthy nor untrustworthy & 91 & \pcb{31.3}\% \\
\rowcolor[HTML]{F2F2F2} Trustworthy & 100 & \pcb{34.4}\% \\
Very trustworthy & 16 & \pcb{5.50}\% \\
\midrule

{\cellcolor{white}}\textbf{How trustworthy or untrustworthy do you rate the news reported by the following media organizations? - The Metro} & {\cellcolor{white}} & {\cellcolor{white}} \\

\midrule
Don't know & 18 & \pcb{6.19}\% \\
\rowcolor[HTML]{F2F2F2}Very untrustworthy & 39 & \pcb{13.4}\% \\
Untrustworthy & 64 & \pcb{22.0}\% \\
\rowcolor[HTML]{F2F2F2}Neither trustworthy nor untrustworthy & 111 & \pcb{38.1}\% \\
Trustworthy & 54 & \pcb{18.6}\% \\
\rowcolor[HTML]{F2F2F2} Very trustworthy & 5 & \pcb{1.72}\% \\
\midrule

{\cellcolor{white}}\textbf{How trustworthy or untrustworthy do you rate the news reported by the following media organizations? - GB News} & {\cellcolor{white}} & {\cellcolor{white}} \\

\midrule
{\cellcolor{white}} Don't know & 17 & \pcb{5.84}\% \\
    \rowcolor[HTML]{F2F2F2} Very untrustworthy & 91 & \pcb{31.3}\% \\
    {\cellcolor{white}} Untrustworthy & 72 & \pcb{24.7}\% \\
    \rowcolor[HTML]{F2F2F2} Neither trustworthy nor untrustworthy & 69 & \pcb{23.7}\% \\
    {\cellcolor{white}} Trustworthy & 36 & \pcb{12.4}\% \\
    \rowcolor[HTML]{F2F2F2} Very trustworthy & 6 & \pcb{2.06}\% \\
\midrule

{\cellcolor{white}}\textbf{How trustworthy or untrustworthy do you rate the news reported by the following media organizations? - The Daily Mail} & {\cellcolor{white}} & {\cellcolor{white}} \\

\midrule
        Don't know & 6 & \pcb{2.06}\% \\
        \rowcolor[HTML]{F2F2F2} Very untrustworthy & 125 & \pcb{43.0}\% \\
        Untrustworthy & 81 & \pcb{27.8}\% \\
        \rowcolor[HTML]{F2F2F2} Neither trustworthy nor untrustworthy & 47 & \pcb{16.2}\% \\
        Trustworthy & 27 & \pcb{9.28}\% \\
        \rowcolor[HTML]{F2F2F2} Very trustworthy & 5 & \pcb{1.72}\% \\
\midrule

{\cellcolor{white}}\textbf{How trustworthy or untrustworthy do you rate the news reported by the following media organizations? - The Mirror} & {\cellcolor{white}} & {\cellcolor{white}} \\

\midrule
{\cellcolor{white}} Don't know & 8 & \pcb{2.75}\% \\
\rowcolor[HTML]{F2F2F2} Very untrustworthy & 101 & \pcb{34.7}\% \\
Untrustworthy & 89 & \pcb{30.6}\% \\
\rowcolor[HTML]{F2F2F2} Neither trustworthy nor untrustworthy & 57 & \pcb{19.6}\% \\
Trustworthy & 33 & \pcb{11.3}\% \\
\rowcolor[HTML]{F2F2F2} Very trustworthy & 3 & \pcb{1.03}\% \\
\midrule

{\cellcolor{white}}\textbf{Cognitive Reflection Test - Number of Correct Answers} & {\cellcolor{white}}  & {\cellcolor{white}}  \\
\midrule
0 & 58 & \pcb{19.9}\% \\
\rowcolor[HTML]{F2F2F2}1 & 44 & \pcb{15.1}\% \\
2 & 67 & \pcb{23.0}\% \\
\rowcolor[HTML]{F2F2F2}3 & 122 & \pcb{41.9}\% \\
\midrule

{\cellcolor{white}}\textbf{Extraversion} & {\cellcolor{white}}  & {\cellcolor{white}}  \\
\midrule
\rowcolor[HTML]{F2F2F2}1 & 37 & \pcb{12.7}\% \\
1.5 & 24 & \pcb{8.25}\% \\
\rowcolor[HTML]{F2F2F2}2 & 43 & \pcb{14.8}\% \\
2.5 & 49 & \pcb{16.8}\% \\
\rowcolor[HTML]{F2F2F2}3 & 59 & \pcb{20.3}\% \\
3.5 & 17 & \pcb{5.84}\% \\
\rowcolor[HTML]{F2F2F2}4 & 34 & \pcb{11.7}\% \\
4.5 & 16 & \pcb{5.50}\% \\
\rowcolor[HTML]{F2F2F2}5 & 12 & \pcb{4.12}\% \\
\midrule

{\cellcolor{white}}\textbf{Agreeableness} & {\cellcolor{white}} & {\cellcolor{white}}  \\
\midrule
\rowcolor[HTML]{F2F2F2}1 & 6 & \pcb{2.06}\% \\
1.5 & 8 & \pcb{2.75}\% \\
\rowcolor[HTML]{F2F2F2}2 & 23 & \pcb{7.90}\% \\
2.5 & 24 & \pcb{8.25}\% \\
\rowcolor[HTML]{F2F2F2}3 & 59 & \pcb{20.3}\% \\
3.5 & 59 & \pcb{20.3}\% \\
\rowcolor[HTML]{F2F2F2}4 & 42 & \pcb{14.4}\% \\
4.5 & 44 & \pcb{15.1}\% \\
\rowcolor[HTML]{F2F2F2}5 & 26 & \pcb{8.93}\% \\
\midrule

{\cellcolor{white}}\textbf{Conscientiousness} & {\cellcolor{white}}  & {\cellcolor{white}} \\
\midrule
\rowcolor[HTML]{F2F2F2}1.5 & 4 & \pcb{1.37}\% \\
2 & 15 & \pcb{5.15}\% \\
\rowcolor[HTML]{F2F2F2}2.5 & 12 & \pcb{4.12}\% \\
3 & 41 & \pcb{14.1}\% \\
\rowcolor[HTML]{F2F2F2}3.5 & 54 & \pcb{18.6}\% \\
4 & 59 & \pcb{20.3}\% \\
\rowcolor[HTML]{F2F2F2}4.5 & 44 & \pcb{15.1}\% \\
5 & 62 & \pcb{21.3}\% \\
\midrule

{\cellcolor{white}}\textbf{Neuroticism} & {\cellcolor{white}} & {\cellcolor{white}}\\
\midrule
\rowcolor[HTML]{F2F2F2}1 & 30 & \pcb{10.3}\% \\
1.5 & 23 & \pcb{7.90}\% \\
\rowcolor[HTML]{F2F2F2}2 & 34 & \pcb{11.7}\% \\
2.5 & 36 & \pcb{12.4}\% \\
\rowcolor[HTML]{F2F2F2}3 & 52 & \pcb{17.9}\% \\
3.5 & 27 & \pcb{9.28}\% \\
\rowcolor[HTML]{F2F2F2}4 & 43 & \pcb{14.8}\% \\
4.5 & 28 & \pcb{9.62}\% \\
\rowcolor[HTML]{F2F2F2}5 & 18 & \pcb{6.19}\% \\
\midrule

{\cellcolor{white}}\textbf{Openness} & {\cellcolor{white}} & {\cellcolor{white}} \\
\midrule
\rowcolor[HTML]{F2F2F2}1 & 1 & \pcb{0.344}\% \\
1.5 & 5 & \pcb{1.72}\% \\
\rowcolor[HTML]{F2F2F2}2 & 18 & \pcb{6.19}\% \\
2.5 & 24 & \pcb{8.25}\% \\
\rowcolor[HTML]{F2F2F2}3 & 51 & \pcb{17.5}\% \\
3.5 & 61 & \pcb{21.0}\% \\
\rowcolor[HTML]{F2F2F2}4 & 47 & \pcb{16.2}\% \\
4.5 & 43 & \pcb{14.8}\% \\
\rowcolor[HTML]{F2F2F2}5 & 41 & \pcb{14.1}\% \\
\midrule

{\cellcolor{white}}\textbf{Perth Emotional Reactivity Scale - Positive Activation: Please score the following statements according to how
much they apply or do not apply to you. - I tend to get happy very easily.} & {\cellcolor{white}} & {\cellcolor{white}} \\
\midrule
\rowcolor[HTML]{F2F2F2} Very unlike me & 24 & \pcb{8.25}\% \\
Somewhat unlike me & 68 & \pcb{23.4}\% \\
\rowcolor[HTML]{F2F2F2} Neither like or unlike me & 67 & \pcb{23.0}\% \\
Somewhat like me & 104 & \pcb{35.7}\% \\
\rowcolor[HTML]{F2F2F2} Very like me & 28 & \pcb{9.62}\% \\
\midrule

{\cellcolor{white}}\textbf{Perth Emotional Reactivity Scale - Positive Intensity: Please score the following statements according to how
much they apply or do not apply to you. - I experience positive mood very strongly.} & {\cellcolor{white}} & {\cellcolor{white}} \\
\midrule
\rowcolor[HTML]{F2F2F2} Very unlike me & 14 & \pcb{4.81}\% \\
Somewhat unlike me & 48 & \pcb{16.5}\% \\
\rowcolor[HTML]{F2F2F2} Neither like or unlike me & 63 & \pcb{21.6}\% \\
Somewhat like me & 130 & \pcb{44.7}\% \\
\rowcolor[HTML]{F2F2F2} Very like me & 36 & \pcb{12.4}\% \\
\midrule

{\cellcolor{white}}\textbf{Perth Emotional Reactivity Scale - Negative Activation: Please score the following statements according to how
much they apply or do not apply to you. - I tend to get disappointed very easily.} & {\cellcolor{white}}& {\cellcolor{white}} \\
    \midrule
    \rowcolor[HTML]{F2F2F2} Somewhat like me & 101 & \pcb{34.7}\% \\
    Somewhat unlike me & 78 & \pcb{26.8}\% \\
    \rowcolor[HTML]{F2F2F2} Neither like or unlike me & 47 & \pcb{16.2}\% \\
    Very like me & 40 & \pcb{13.7}\% \\
    \rowcolor[HTML]{F2F2F2} Very unlike me & 25 & \pcb{8.59}\% \\
    \midrule

{\cellcolor{white}}\textbf{Perth Emotional Reactivity Scale - Negative Intensity: Please score the following statements according to how
much they apply or do not apply to you. - My negative feelings feel very intense.} & {\cellcolor{white}} & {\cellcolor{white}} \\
\midrule
\rowcolor[HTML]{F2F2F2} Somewhat like me & 94 & \pcb{32.3}\% \\
Somewhat unlike me & 60 & \pcb{20.6}\% \\
\rowcolor[HTML]{F2F2F2} Very like me & 55 & \pcb{18.9}\% \\
Neither like or unlike me & 49 & \pcb{16.8}\% \\
\rowcolor[HTML]{F2F2F2} Very unlike me & 33 & \pcb{11.3}\% \\
\midrule

{\cellcolor{white}}\textbf{Positive and Negative Affect Schedule - Net Positive Score} & {\cellcolor{white}} \textbf{Count} & {\cellcolor{white}} \textbf{Percentage} \\
\hline
\rowcolor[HTML]{F2F2F2} < -10 & 1 & \pcb{0.344}\% \\
-10 to -5 & 1 & \pcb{0.344}\% \\
\rowcolor[HTML]{F2F2F2} -5 to 0 & 23 & \pcb{7.90}\% \\
0 to 5 & 77 & \pcb{26.5}\% \\
\rowcolor[HTML]{F2F2F2} 5 to 10 & 112 & \pcb{38.5}\% \\
> 10 & 77 & \pcb{26.5}\% \\

\end{longtable}
}
\clearpage
\twocolumn

\begin{table*}[!htbp]
\centering
\begin{tabular}{l*{6}{c}}
\toprule
& \multicolumn{2}{c}{Image} & \multicolumn{2}{c}{Text-only} & \multicolumn{2}{c}{Statistics} \\
\cmidrule(lr){2-3} \cmidrule(lr){4-5} \cmidrule(lr){6-7}
Dimension & $M$ ($SD$) & $Mdn$ & $M$ ($SD$) & $Mdn$ & $U$ & $p$-value \\
\midrule
Valence & $3.03$ ($1.40$) & $3.0$ & $3.35$ ($1.36$) & $3.0$ & $17{,}272.5$ & $.016^{*}$ \\
Arousal & $4.02$ ($1.42$) & $4.0$ & $4.09$ ($1.53$) & $4.0$ & $19{,}366.0$ & $.574$ \\
Dominance & $3.66$ ($1.04$) & $4.0$ & $3.90$ ($1.19$) & $4.0$ & $17{,}529.5$ & $.019^{*}$ \\
\bottomrule
\end{tabular}
\caption{Comparison of affective response between image and text-only conditions.* denotes p < .05.}
\label{tab:modality_comparison}
\end{table*}

\subsection{Pilot Study: Stimulus Modality Selection}
\label{sec:pre_study}
We conducted a pilot study to determine whether to use full Facebook news post screenshots or text-only headlines as stimuli. This decision involves several trade-offs. Text-only stimuli are simpler to process with current language models and sufficient for many breaking news posts that use generic images. However, full screenshots offer greater ecological validity, as social media users typically encounter both text and images simultaneously. Prior research suggests that visual stimuli are processed more rapidly than text~\cite{Azizian2006} and are more memorable~\cite{SHEPARD1967156}, though current open-source vision-language models still face significant performance and robustness challenges~\cite{li-etal-2024-topviewrs, sterz2024dare}.

To empirically inform this decision, we collect annotations from 40 UK-based participants (20 per condition) for 10 paired news posts from March 2024, present either as text-only headlines or full screenshots. Participants rated valence, arousal, and dominance (VAD) and provide discrete emotion categories.

\begin{figure}[htbp]
\centering
\includegraphics[width=0.47\textwidth]{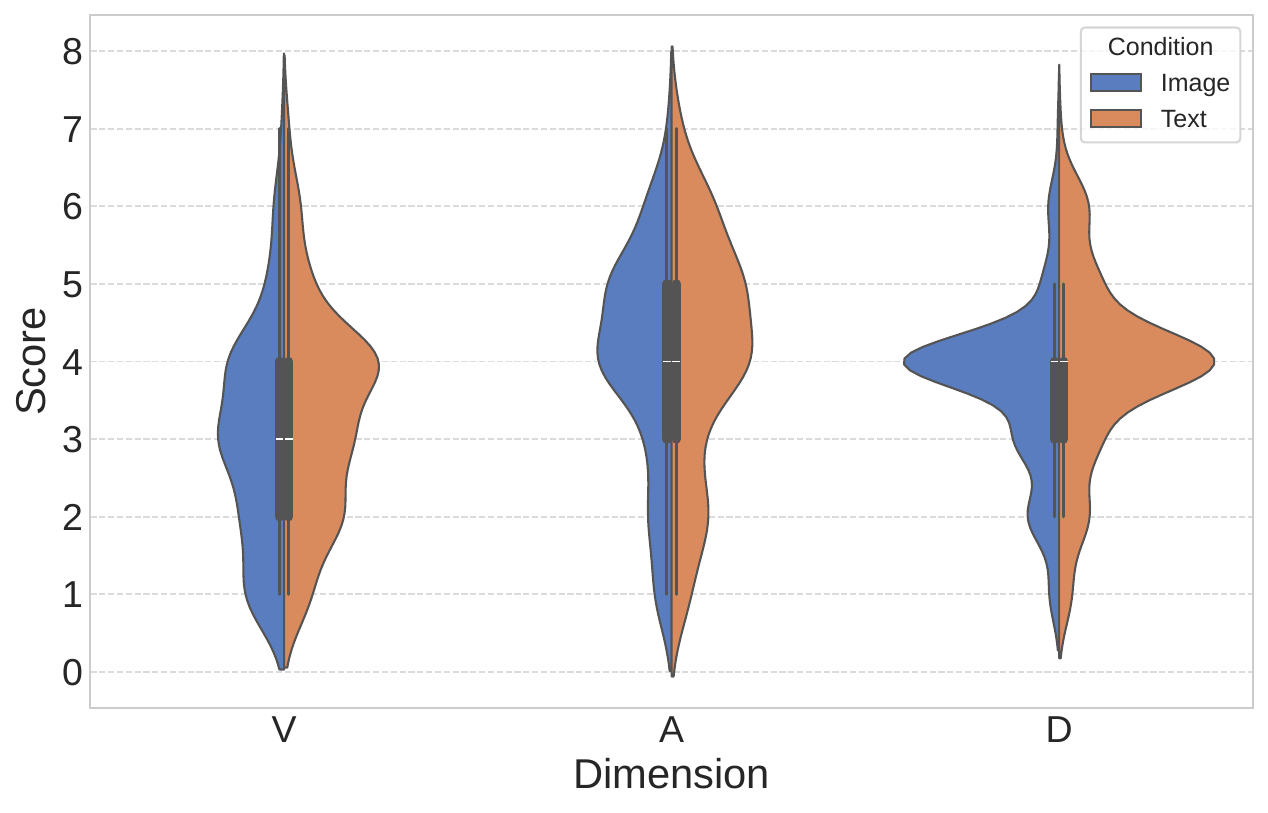}
\caption{Distribution of VAD scores across modality conditions}
\label{fig:vad_dist}
\end{figure}

\begin{figure}[htbp]
\centering
\includegraphics[width=0.47\textwidth]{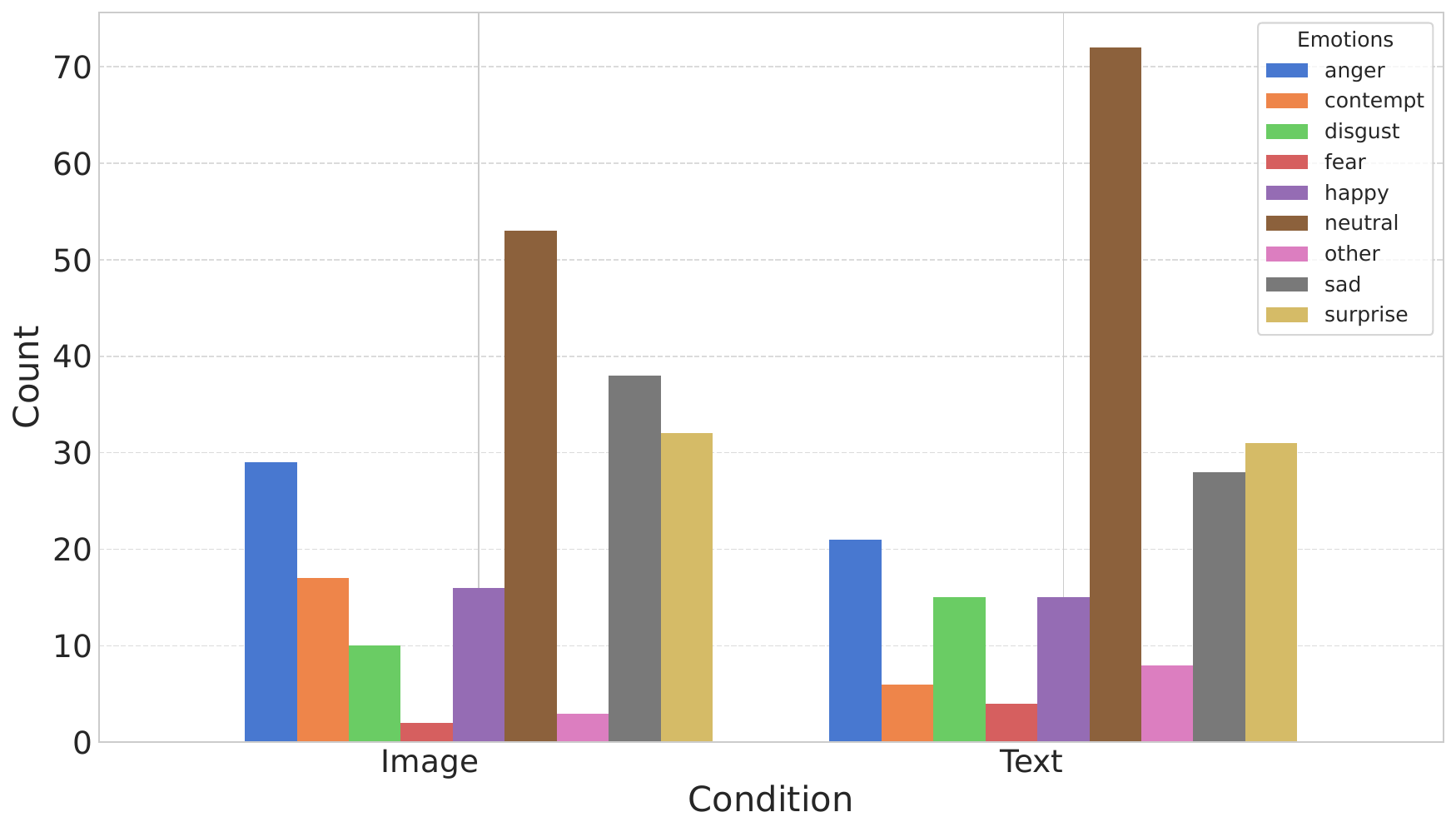}
\caption{Distribution of discrete emotions across modality conditions}
\label{fig:emotion_dist}
\end{figure}

We then analyze the aggregated ratings across all 10 posts for each condition. Table~\ref{tab:modality_comparison} presents the descriptive and inferential statistics for the dimensional emotions (VAD). Mann-Whitney U tests indicate significant differences in valence $(p = 0.016)$ and dominance $(p = 0.019)$, though with small effect sizes  $(RBC \text{ ranging from } -0.032 \text{ to } -0.136)$. For discrete emotions, a chi-square test indicates marginally significant differences in emotion distribution between conditions $(\chi^2 = 14.93, p = 0.060)$. We additionally visually show the distribution of VAD and discrete ratings in Figures~\ref{fig:vad_dist} and~\ref{fig:emotion_dist}.

The distribution patterns of VAD and discrete ratings are visualized in Figures \ref{fig:vad_dist} and \ref{fig:emotion_dist}. While VAD distributions remain broadly similar across conditions, the image condition elicits more negative valence ratings and more neutral dominance ratings. The differences in discrete emotion ratings are more noticeable, with substantially fewer neutral emotions reported in the image condition. We interpret this as evidence that images help disambiguate emotional content - since the image condition includes both visual and textual information, it may provide richer context for emotional interpretation.

Based on these findings and theoretical considerations, we decide to use full screenshots for our main study. This choice is driven by observed differences in emotional annotations, the ecological validity of multimodal news consumption on social media, and the additional contextual information provided by images. While current vision-language models face technical limitations, we anticipate rapid advancement in multimodal processing capabilities and prioritize capturing more naturalistic news consumption experiences over immediate computational convenience.

\begin{table*}
\centering
\centering
\resizebox{\ifdim\width>\linewidth\linewidth\else\width\fi}{!}{
\begin{tabular}[htbp]{>{\raggedright\arraybackslash}p{3cm}rrrrrrrl}
\toprule
Page Name & Mean V & SD V & Mean A & SD A & Mean D & SD D & Count & Discrete \%\\
\midrule
\cellcolor{gray!10}{BBC News} & \cellcolor{gray!10}{3.45} & \cellcolor{gray!10}{1.08} & \cellcolor{gray!10}{4.42} & \cellcolor{gray!10}{1.14} & \cellcolor{gray!10}{3.72} & \cellcolor{gray!10}{1.01} & \cellcolor{gray!10}{498} & \cellcolor{gray!10}{sad: 24.6\%, neutral: 22.3\%}\\
The Independent & 3.49 & 0.99 & 4.28 & 1.12 & 3.82 & 0.85 & 339 & neutral: 34.6\%, sad: 17.5\%\\
\cellcolor{gray!10}{Daily Mail} & \cellcolor{gray!10}{3.09} & \cellcolor{gray!10}{0.97} & \cellcolor{gray!10}{4.40} & \cellcolor{gray!10}{1.21} & \cellcolor{gray!10}{3.61} & \cellcolor{gray!10}{1.00} & \cellcolor{gray!10}{305} & \cellcolor{gray!10}{sad: 28.4\%, neutral: 26.7\%}\\
The Mirror & 3.59 & 0.97 & 4.03 & 1.19 & 3.85 & 0.83 & 240 & neutral: 40.1\%, sad: 18.7\%\\
\cellcolor{gray!10}{Metro} & \cellcolor{gray!10}{3.44} & \cellcolor{gray!10}{0.98} & \cellcolor{gray!10}{4.19} & \cellcolor{gray!10}{1.19} & \cellcolor{gray!10}{3.77} & \cellcolor{gray!10}{0.91} & \cellcolor{gray!10}{213} & \cellcolor{gray!10}{neutral: 35.6\%, sad: 19\%}\\
\addlinespace
The Sun & 3.56 & 1.03 & 4.07 & 1.27 & 3.88 & 0.87 & 213 & neutral: 35.9\%, sad: 17\%\\
\cellcolor{gray!10}{The Telegraph} & \cellcolor{gray!10}{3.37} & \cellcolor{gray!10}{1.15} & \cellcolor{gray!10}{4.39} & \cellcolor{gray!10}{1.12} & \cellcolor{gray!10}{3.79} & \cellcolor{gray!10}{1.04} & \cellcolor{gray!10}{190} & \cellcolor{gray!10}{neutral: 29\%, sad: 16.2\%}\\
Daily Express & 3.70 & 0.91 & 3.85 & 1.17 & 3.89 & 0.70 & 141 & neutral: 49.5\%, sad: 13.3\%\\
\cellcolor{gray!10}{The Guardian} & \cellcolor{gray!10}{3.68} & \cellcolor{gray!10}{1.06} & \cellcolor{gray!10}{4.21} & \cellcolor{gray!10}{1.19} & \cellcolor{gray!10}{3.88} & \cellcolor{gray!10}{0.96} & \cellcolor{gray!10}{136} & \cellcolor{gray!10}{neutral: 30.5\%, sad: 20.7\%}\\
The Economist & 3.61 & 0.99 & 4.35 & 1.08 & 3.74 & 0.93 & 126 & neutral: 39.3\%, happy: 11\%\\
\addlinespace
\cellcolor{gray!10}{Daily Star} & \cellcolor{gray!10}{3.67} & \cellcolor{gray!10}{1.10} & \cellcolor{gray!10}{4.12} & \cellcolor{gray!10}{1.16} & \cellcolor{gray!10}{4.01} & \cellcolor{gray!10}{0.79} & \cellcolor{gray!10}{109} & \cellcolor{gray!10}{neutral: 39.5\%, happy: 13.3\%}\\
The i Paper & 3.32 & 1.17 & 4.42 & 1.10 & 3.72 & 1.09 & 89 & neutral: 26.3\%, sad: 19.1\%\\
\cellcolor{gray!10}{ITV News} & \cellcolor{gray!10}{3.35} & \cellcolor{gray!10}{1.13} & \cellcolor{gray!10}{4.45} & \cellcolor{gray!10}{1.10} & \cellcolor{gray!10}{3.70} & \cellcolor{gray!10}{1.10} & \cellcolor{gray!10}{69} & \cellcolor{gray!10}{sad: 22.5\%, neutral: 21.9\%}\\
The Times and The Sunday Times & 3.30 & 1.23 & 4.55 & 1.17 & 3.71 & 1.15 & 50 & neutral: 23.8\%, anger: 17.2\%\\
\cellcolor{gray!10}{LADbible} & \cellcolor{gray!10}{3.67} & \cellcolor{gray!10}{0.98} & \cellcolor{gray!10}{4.46} & \cellcolor{gray!10}{1.18} & \cellcolor{gray!10}{3.81} & \cellcolor{gray!10}{1.03} & \cellcolor{gray!10}{49} & \cellcolor{gray!10}{surprise: 24.7\%, neutral: 21.2\%}\\
\addlinespace
Sky News & 2.90 & 1.04 & 4.74 & 1.15 & 3.50 & 1.10 & 41 & sad: 26.6\%, neutral: 19.6\%\\
\cellcolor{gray!10}{GB News} & \cellcolor{gray!10}{3.49} & \cellcolor{gray!10}{1.10} & \cellcolor{gray!10}{4.09} & \cellcolor{gray!10}{1.12} & \cellcolor{gray!10}{3.90} & \cellcolor{gray!10}{0.95} & \cellcolor{gray!10}{40} & \cellcolor{gray!10}{neutral: 31.9\%, sad: 15.2\%}\\
Reuters UK & 3.71 & 1.23 & 4.27 & 1.21 & 3.84 & 1.11 & 24 & neutral: 29.5\%, surprise: 18\%\\
\cellcolor{gray!10}{LBC} & \cellcolor{gray!10}{3.40} & \cellcolor{gray!10}{1.12} & \cellcolor{gray!10}{4.30} & \cellcolor{gray!10}{1.16} & \cellcolor{gray!10}{3.78} & \cellcolor{gray!10}{1.02} & \cellcolor{gray!10}{19} & \cellcolor{gray!10}{sad: 27.6\%, neutral: 19.4\%}\\
Financial Times & 3.95 & 1.18 & 4.28 & 1.20 & 3.87 & 1.14 & 8 & neutral: 39.5\%, surprise: 16.3\%\\
\bottomrule
\end{tabular}}
\caption{Distribution of valence, arousal, and dominance and discrete emotion labels by outlet.}
\label{tab:summary_stats_per_outlet}
\end{table*}

\subsection{Geographic Representation of Annotators}
We present the geographic distribution of annotators across UK postcode areas in Figure \ref{fig:geographic_representation}. Our dataset includes annotators from 97 of the 124 postcode areas in the UK, demonstrating broad geographical coverage. To assess the representativeness of our sample, we compute the Pearson correlation coefficient between the number of annotators per postcode area and the corresponding 2011 Census population figures. The resulting correlation of 0.662 indicates a moderate positive relationship between population density and annotator distribution.
To further quantify geographic representation, we calculate a representativeness ratio for each postcode area by dividing the percentage of annotators in each area by the percentage of the UK population in that same area. The mean ratio of 1.26 indicates that most areas are well-represented, often exceeding proportional representation. While there is some variation (standard deviation of 1.05 and median of 0.95), the overall distribution suggests we achieved strong geographic diversity in our sample.

\begin{figure}[!h] %
    \centering
    \includegraphics[width=1\linewidth]{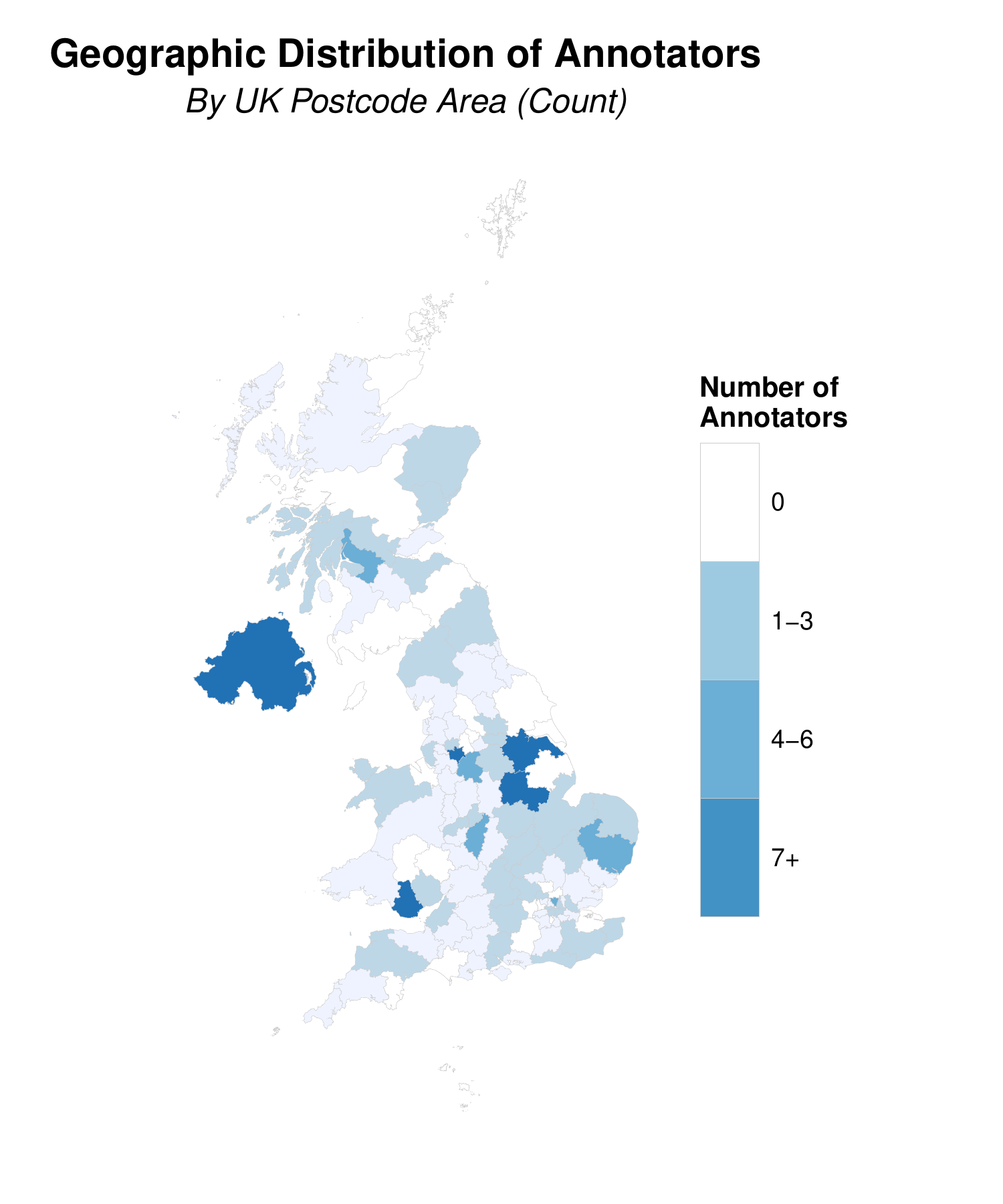}
    \caption{Geographic distribution of annotators across UK postcode areas.}
    \label{fig:geographic_representation}
\end{figure}

\subsection{Annotation Manual}
\label{sec:annotation_manual}
\vspace{-\baselineskip} %
\includepdf[pages=-, pagecommand={},width=0.98\textwidth]{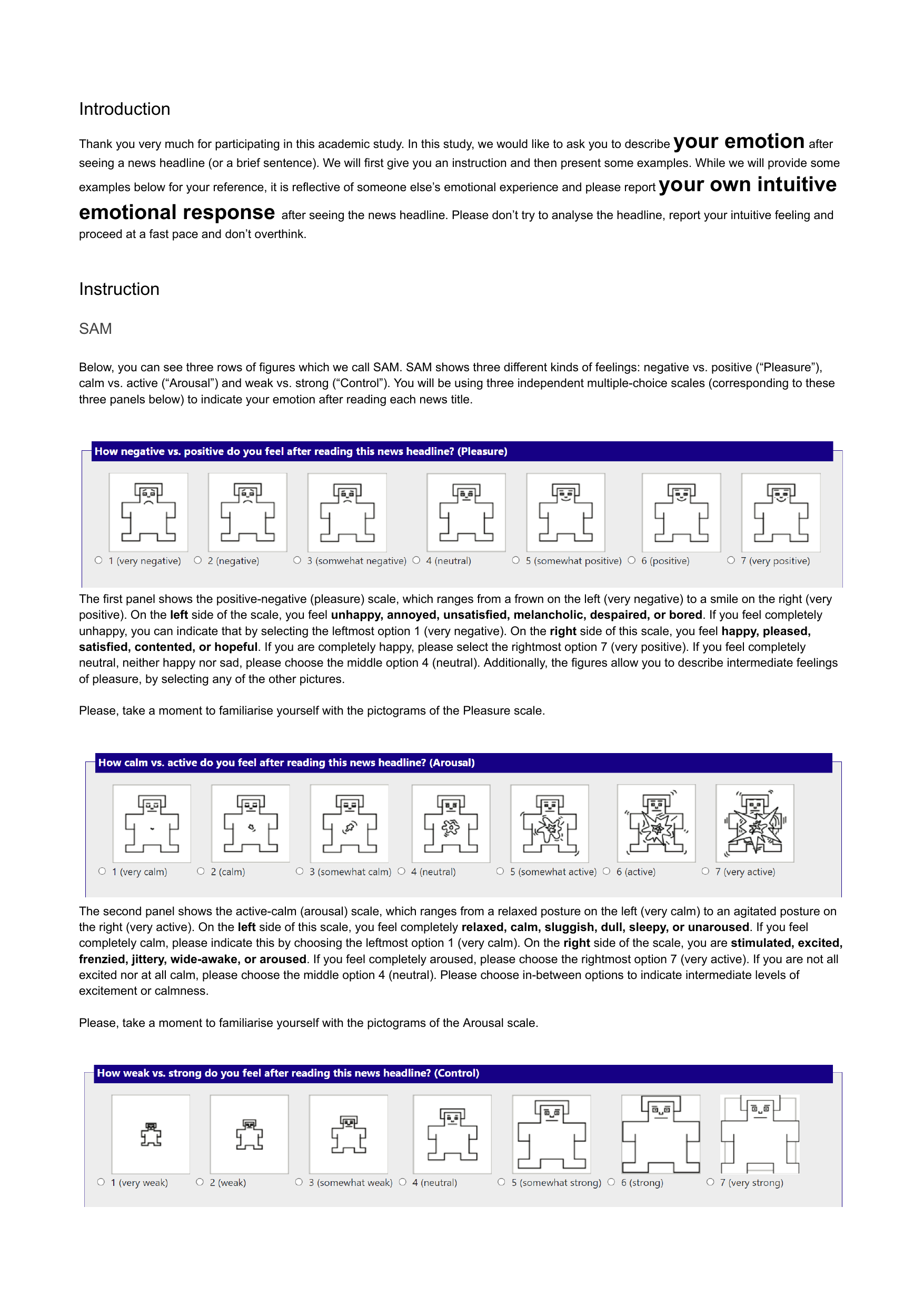}

\subsection{Extended Descriptive Analysis}
\subsubsection{Additional Annotation Distributions Analysis}
\label{sec:auxiliary_variables_analysis}
We show the distributions of the collected annotation variables
in Figure~\ref{fig:all_dist}. In addition to the discussions in Section~\ref{sec:descriptive_analysis}, regarding relevance (Figure~\ref{fig:relevance_dist}), almost half of the annotations (44\%) indicate ``Not at all'' relevant, with only 3.8\% marked as ``extremely relevant.'' For sharing inclination (Figure~\ref{fig:sharing_dist}), the distribution is even more skewed, with 54.5\% of the annotations indicating ``very unlikely'' to share.

The majority of annotations (52.3\%, Figure~\ref{fig:source_dist}) suggest that both the text and image significantly influence emotional reactions to news headlines. In contrast, approximately a third (36.7\%) highlight the text alone as the primary factor. This indicates the importance of considering both the image and the text when modeling affective responses to news headlines on social media, rather than focusing solely on one or the other.

\subsubsection{Outlet-level Analysis}
\label{sec:outlet_level_full}
To examine the traditional distinction between broadsheet and tabloid publications\footnote{We classify The Times, The Telegraph, The Guardian, The Independent, i Paper, and Financial Times as broadsheets, and The Sun, Daily Mail, Daily Express, The Mirror, and Daily Star as tabloids. All other outlets are categorized as ``other''. See~\citet{city15368} for a comparison of broadsheet and tabloid newspapers, both historically and in the present day.}, we conduct Welch's t-tests comparing their affect scores. Interestingly, we find no significant differences in valence ($p=0.83$) or dominance ($p=0.64$) between the two types of outlets. However, there is a marginally significant difference in arousal ($p=0.052$), with broadsheet publications eliciting slightly higher arousal responses ($M=4.34$) compared to tabloids ($M=4.09$). While the digital transformation of news media might have blurred many traditional distinctions between tabloids and broadsheets, these findings suggest that different editorial standards may still influence readers' affective responses, particularly in terms of emotional arousal.

\subsubsection{Relationship Between Arousal and Valence}

\begin{figure}[h] %
    \centering
\includegraphics[width=1\linewidth]{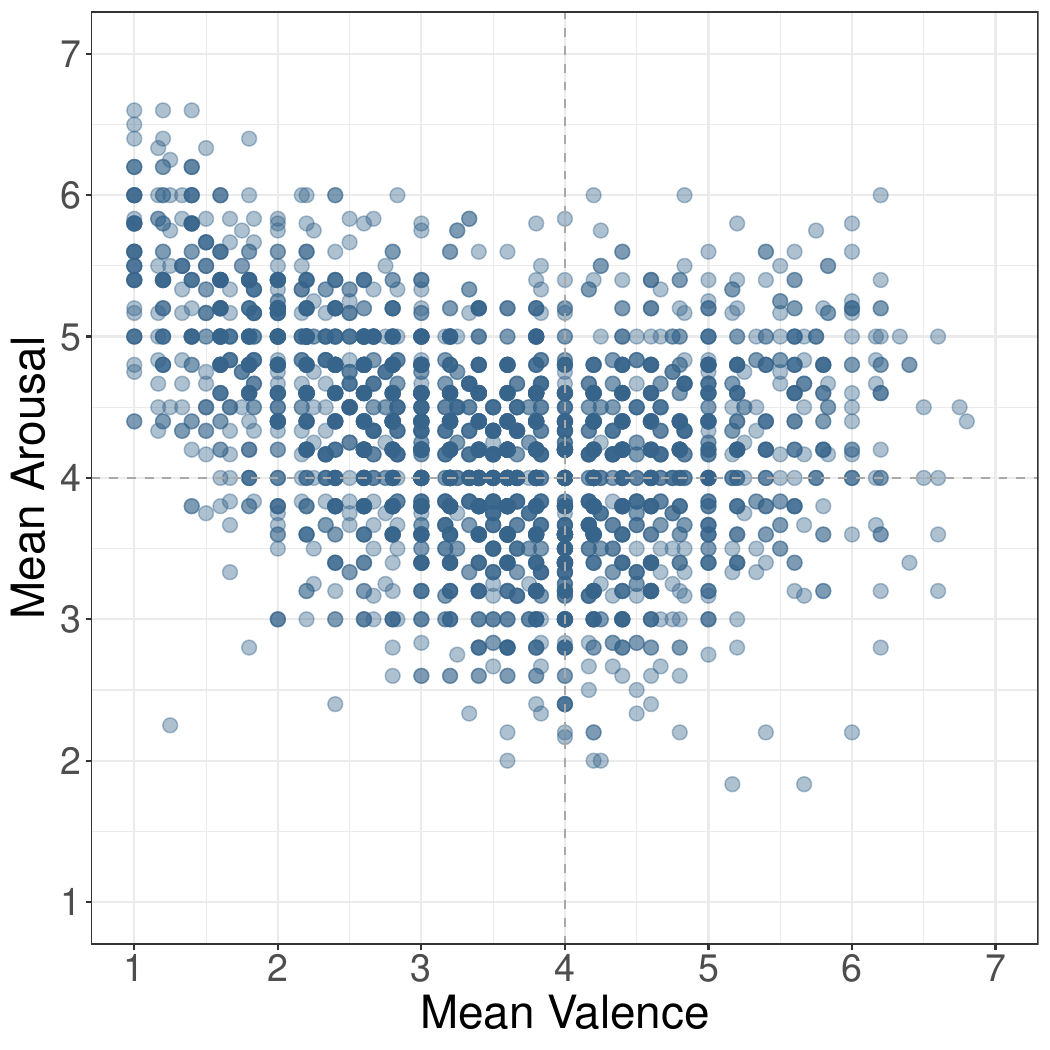} %
    \caption{Distribution of affective responses to news posts in the valence-arousal space. Each point represents the mean arousal and valence ratings for a single headline, with darker regions indicating higher density of overlapping points.}
    \label{fig:arousalvsvalence} %
\end{figure}

We calculate the average valence and arousal for each headline and present the results in Figure~\ref{fig:arousalvsvalence}. Point opacity indicates overlapping points, suggesting a higher density of images. The distribution follows a V-shaped pattern, where arousal levels are high at both low and high extremes of valence, and this pattern aligns with established findings in affective science~\cite{lang1997international,Kurdi2017}. However, our data also present notable deviations. Specifically, we observe a higher concentration of headlines exhibiting elevated arousal levels (above 6) in both the first and second quadrants (low valence/high arousal and high valence/high arousal, respectively). This concentration is particularly pronounced in the second quadrant, characterized by very low valence and very high arousal. We also see a concentration of density along the central region, around arousal $\approx$ 4 and valence $\approx$ 4, with a slight skew towards the upper-left quadrant. Finally, the overall distribution in our dataset encompasses a broader region of the valence-arousal space compared to that of \citet{Kurdi2017}. We hypothesize that this discrepancy arises from the inherently negative nature of news headlines, in contrast to the more emotionally diverse stimuli typically employed in prior studies comprising images of scenes and objects.

\begin{figure*}[!htbp] %
    \centering
    \begin{minipage}[t]{\linewidth}  %
        \centering
        \begin{subfigure}[b]{0.3\textwidth}  %
            \centering
            \includegraphics[width=\textwidth]{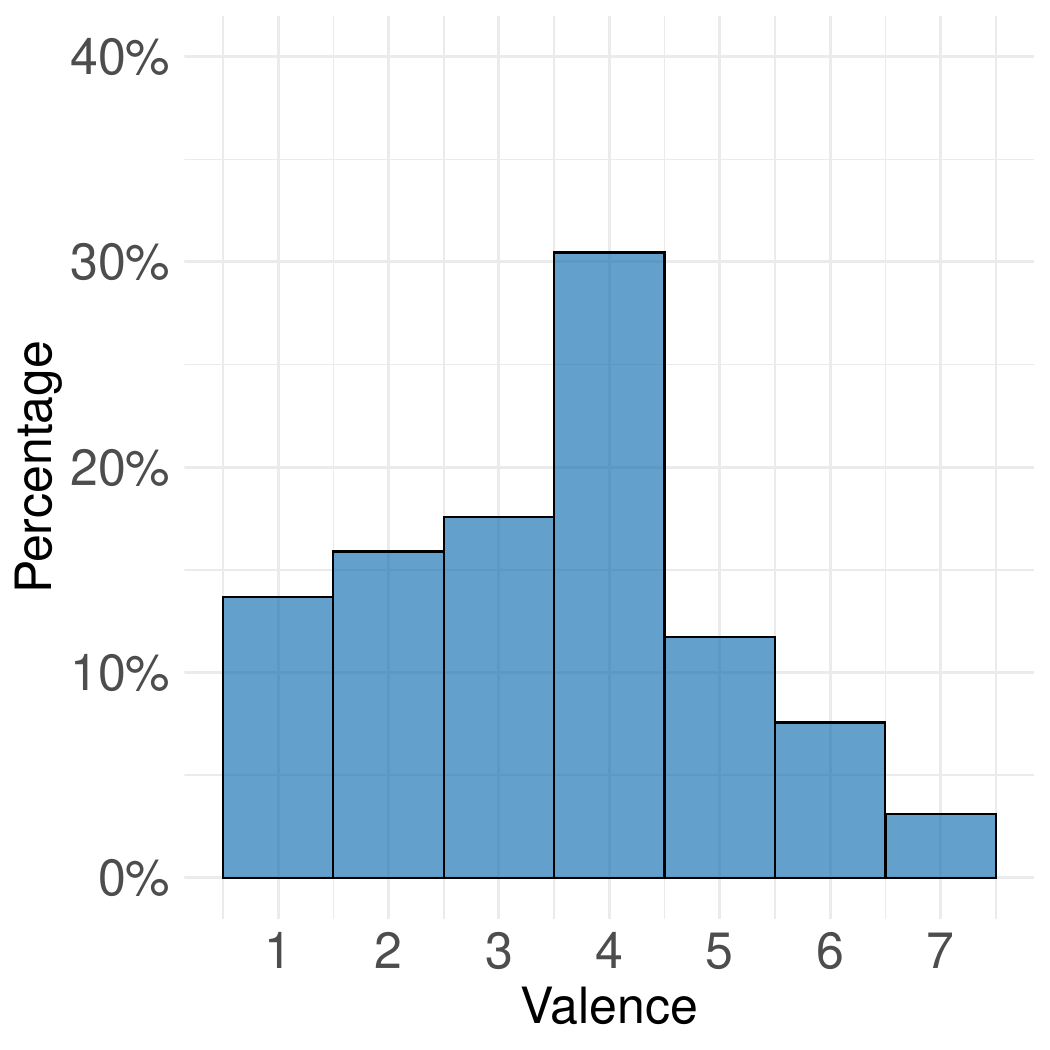}
            \caption{Valence Distribution}
            \label{fig:valence_dist}
        \end{subfigure}
        \hfill
        \begin{subfigure}[b]{0.3\textwidth}
            \centering
            \includegraphics[width=\textwidth]{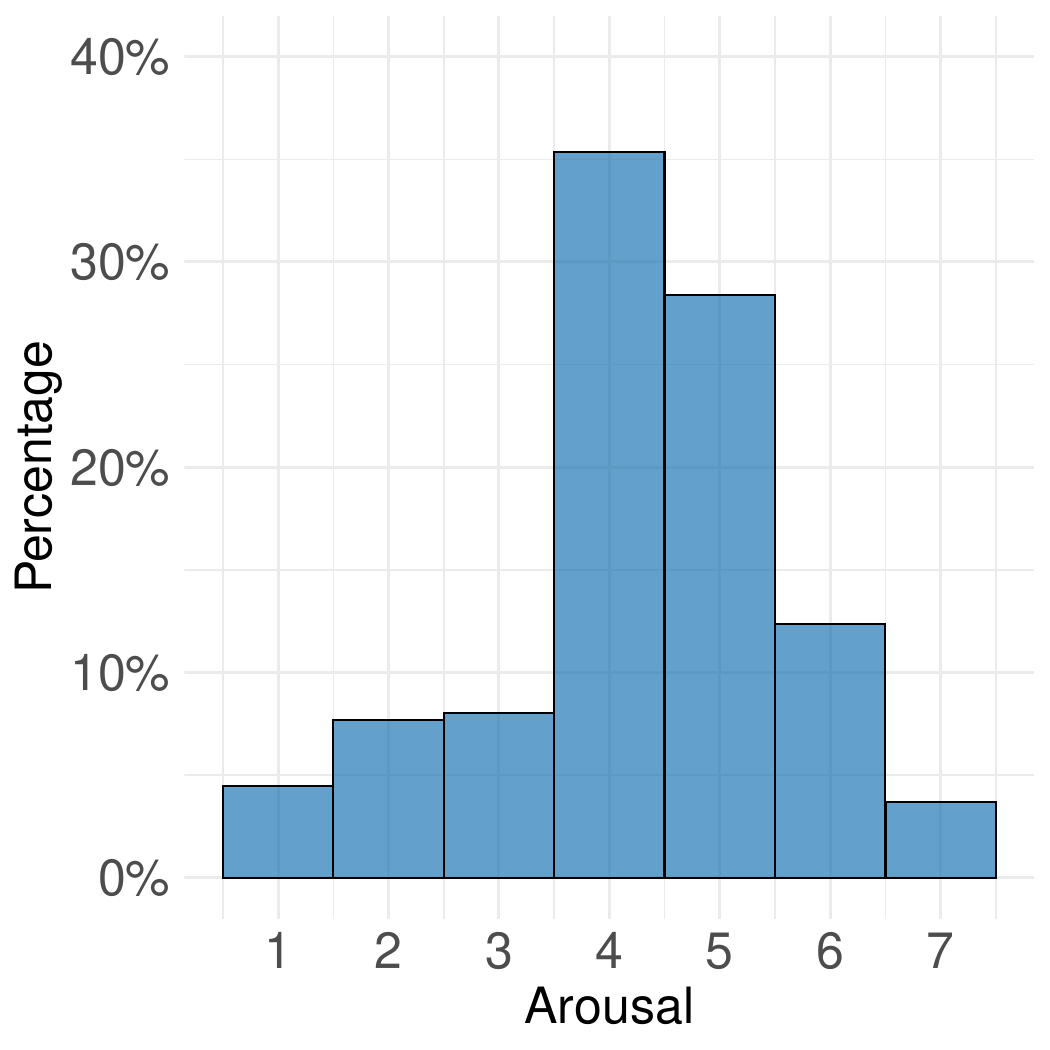}
            \caption{Arousal Distribution}
            \label{fig:arousal_dist}
        \end{subfigure}
        \hfill
        \begin{subfigure}[b]{0.3\textwidth}
            \centering
            \includegraphics[width=\textwidth]{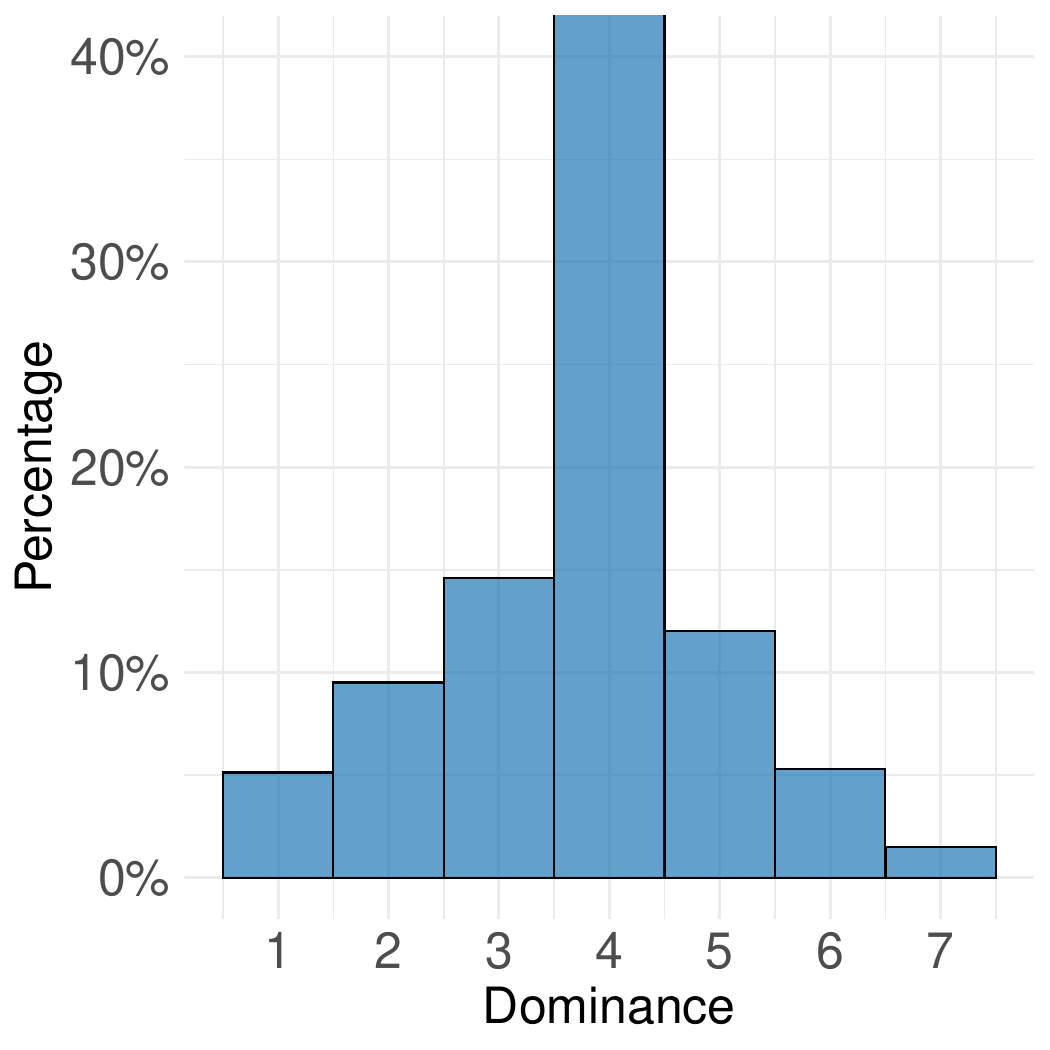}
            \caption{Dominance Distribution}
            \label{fig:dominance_dist}
        \end{subfigure}
    \end{minipage}

    \ContinuedFloat

    \begin{minipage}[t]{\linewidth}
        \centering
        \begin{subfigure}[b]{0.3\textwidth}
            \centering
            \includegraphics[width=\textwidth]{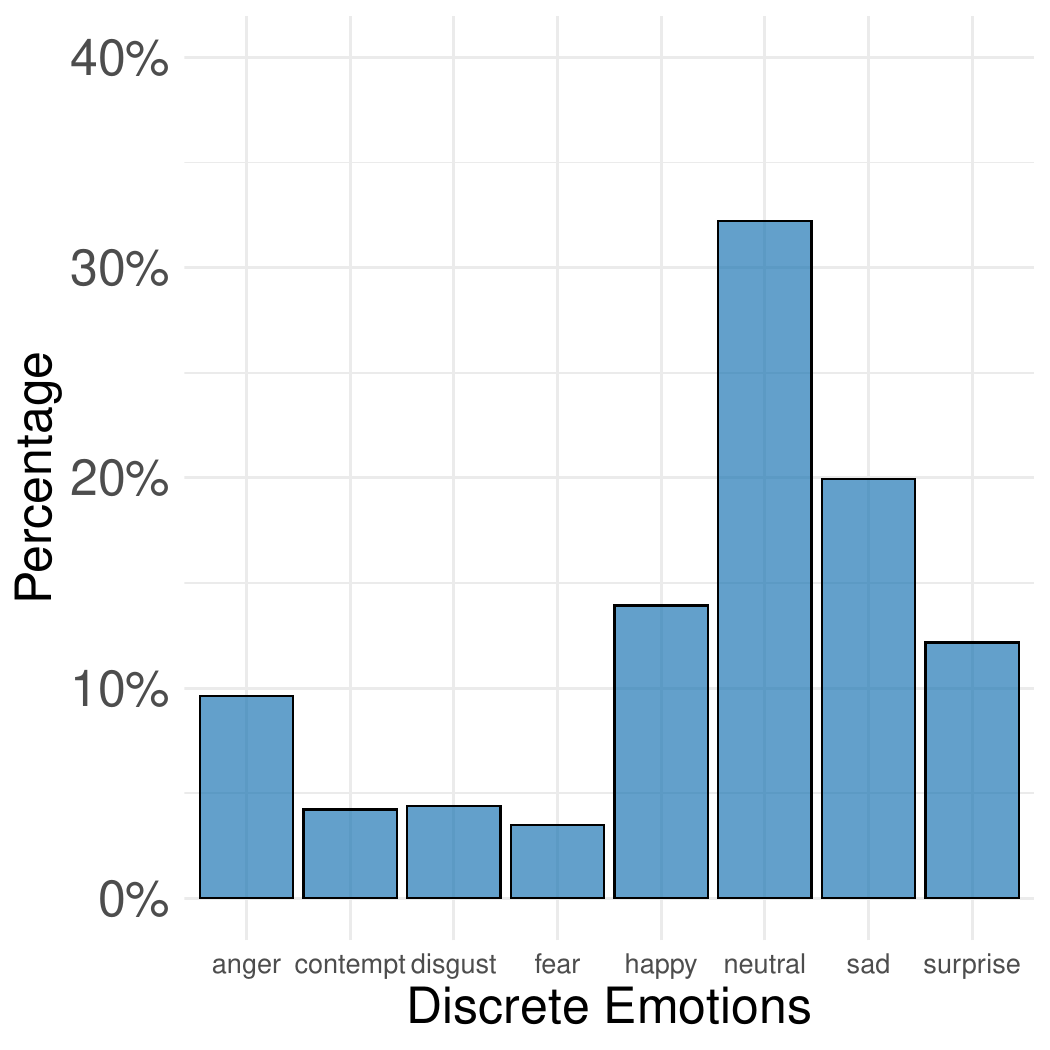}
            \caption{Discrete Distribution}
            \label{fig:discrete_dist}
        \end{subfigure}
        \hfill
        \begin{subfigure}[b]{0.3\textwidth}
            \centering
            \includegraphics[width=\textwidth]{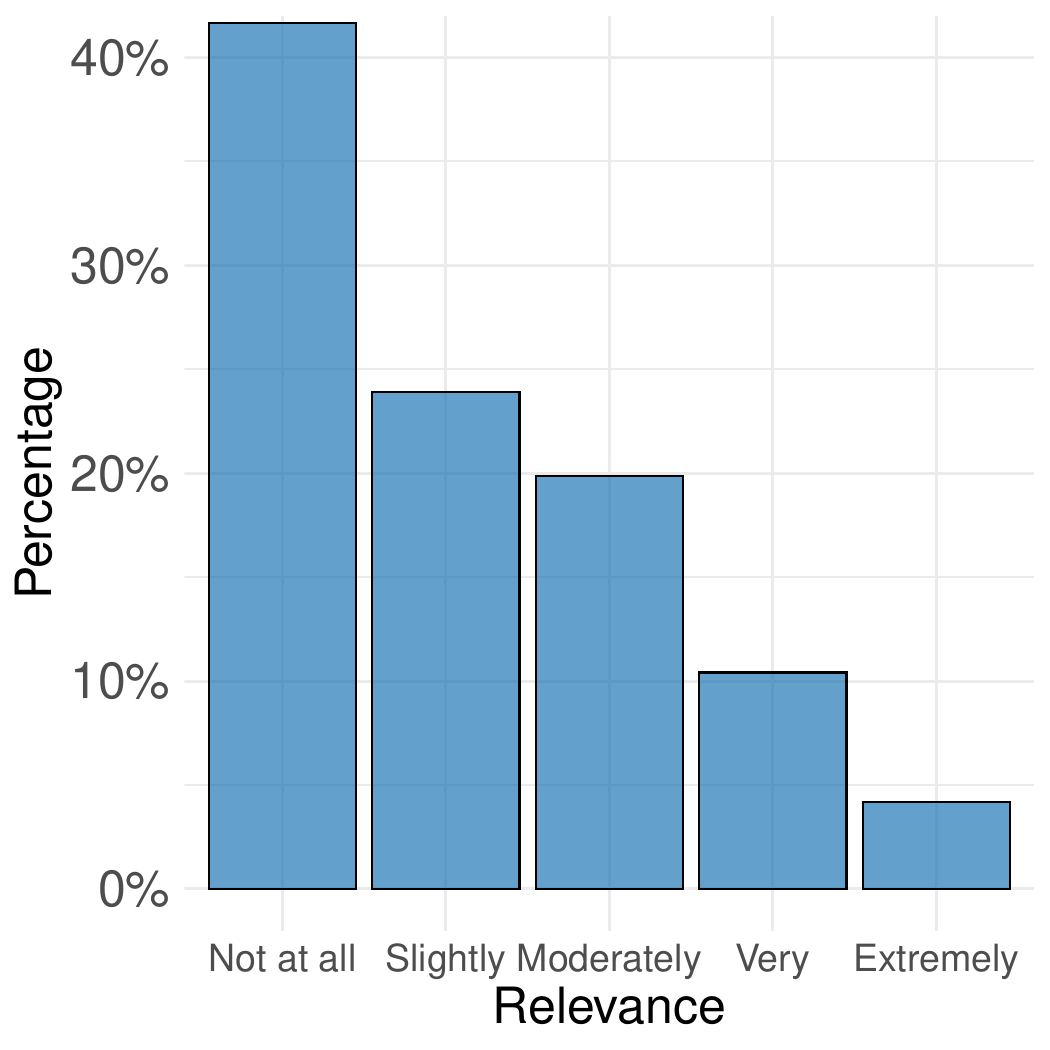}
            \caption{Relevance Distribution}
            \label{fig:relevance_dist}
        \end{subfigure}
        \hfill
        \begin{subfigure}[b]{0.3\textwidth}
            \centering
            \includegraphics[width=\textwidth]{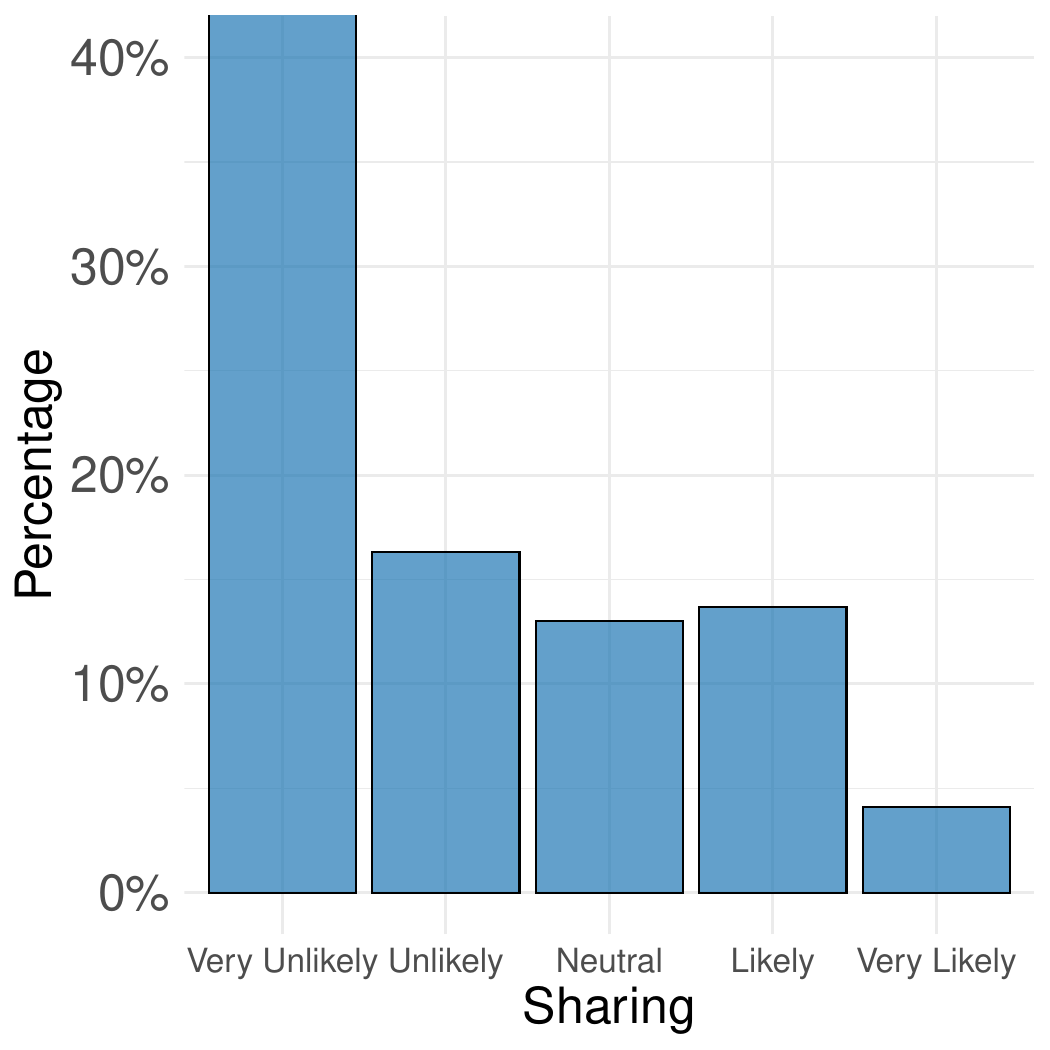}
            \caption{Sharing Distribution}
            \label{fig:sharing_dist}
        \end{subfigure}
    \end{minipage}

    \ContinuedFloat

    \begin{minipage}[t]{\linewidth}
        \centering
        \begin{subfigure}[b]{0.3\textwidth}
            \centering
            \includegraphics[width=\textwidth]{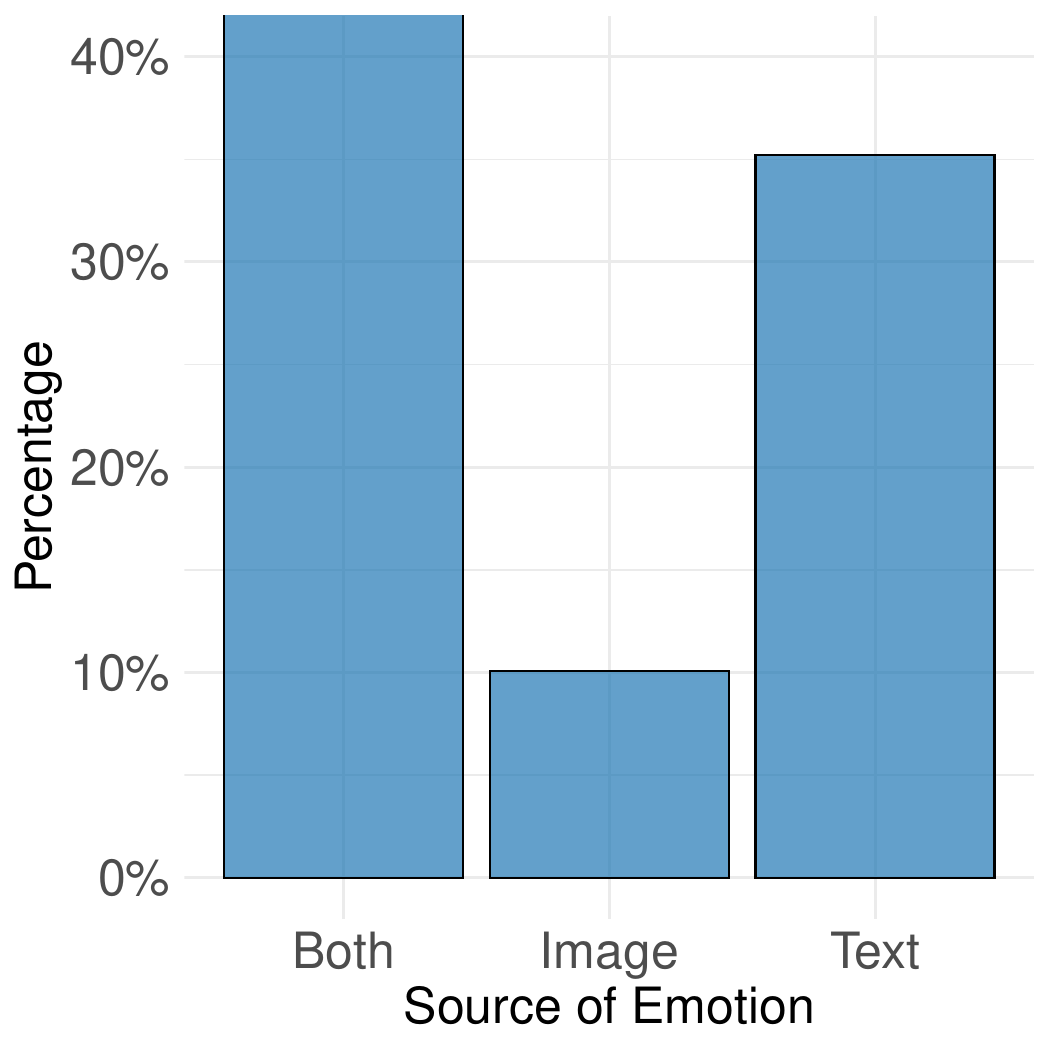}
            \caption{Source Distribution}
            \label{fig:source_dist}
        \end{subfigure}
    \end{minipage}
    \caption{Distribution of Annotations}
    \label{fig:all_dist}

\end{figure*}
\begin{figure*}[htbp] %
    \centering
    \includegraphics[width=0.65\linewidth]{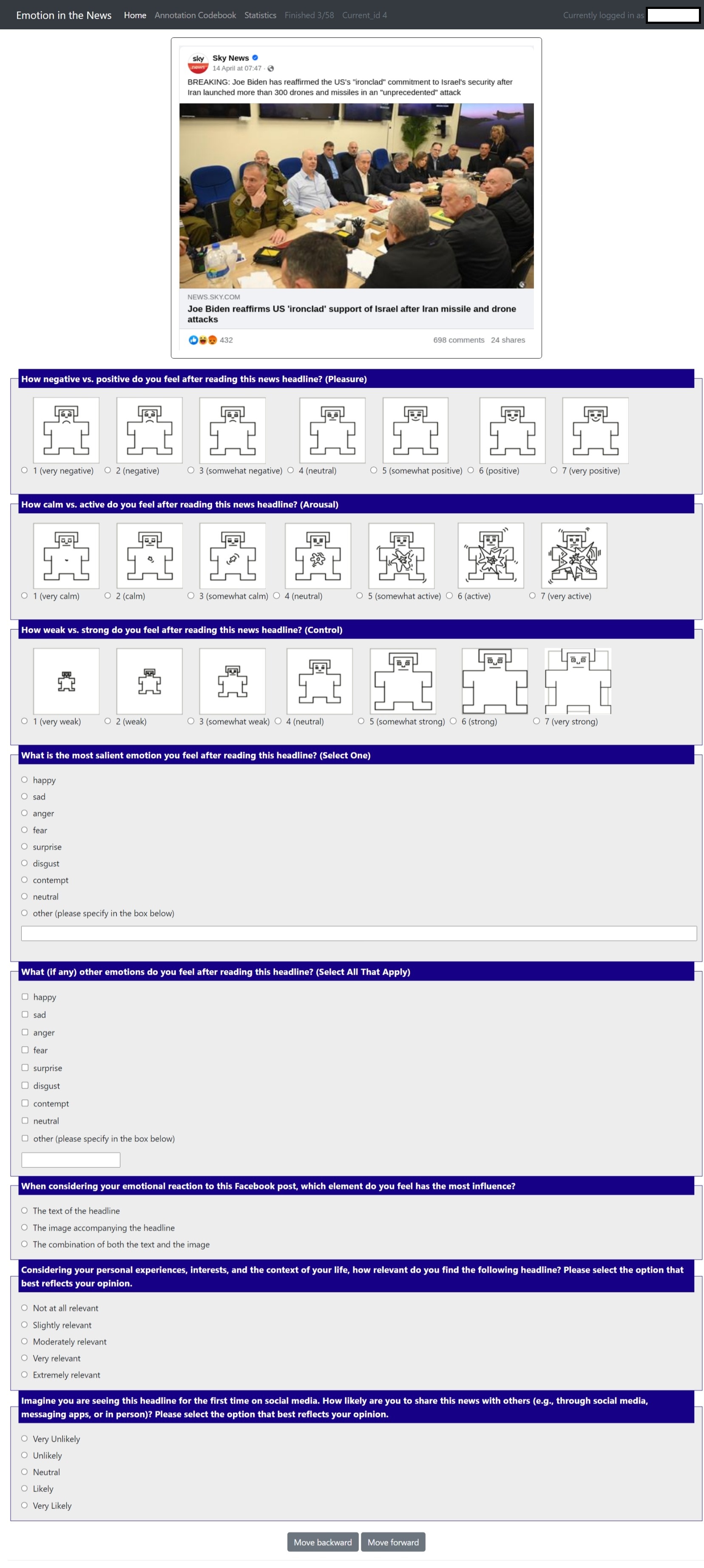} %
    \caption{A screenshot of the annotation interface.}
    \label{fig:screenshot} %
\end{figure*}

\begin{table}[t]
\centering
\begin{tabular}{lc}
\toprule
\textbf{Dimension} & \textbf{Krippendorff's $\alpha$} \\
\midrule
Valence & 0.468 \\
Arousal & 0.145 \\
Dominance & 0.203 \\
Discrete Emotions & 0.202 \\
\midrule
Modality Importance & 0.083\\
Relevance & 0.079\\
Sharing Intent & 0.057\\
\bottomrule
\end{tabular}
\caption{Inter-annotator agreement measured by Krippendorff's $\alpha$ for continuous dimensions (V/A/D), discrete emotions, and auxiliary variables.}
\label{tab:agreement}
\end{table}

\subsection{Inter-annotator agreement}
\label{sec:agreement}
We measure Krippendorff's $\alpha$ for each of the annotated variables and present the results in Table~\ref{tab:agreement}. 
Among the core emotional dimensions, valence shows moderate agreement ($\alpha = 0.468$), while arousal and dominance exhibit lower agreement levels ($\alpha = 0.145$ and $\alpha = 0.203$, respectively). Discrete emotion categories demonstrate comparable levels of agreement. The auxiliary variables—modality importance, relevance, and sharing intent—show particularly low agreement ($\alpha < 0.1$).

These findings align with previous research in emotion and affect annotation. The relatively low inter-annotator agreement is consistent with similar datasets \cite{strapparava2007semeval, Busso2008IEMOCAP, demszky-etal-2020-goemotions, oberlander2020goodnewseveryone}, and the pattern of higher agreement for valence compared to arousal and dominance mirrors observations in prior work \cite{Busso2008IEMOCAP,buechel-hahn-2017-emobank}. These low agreement levels highlight a crucial insight: emotional responses to news content are inherently subjective and individualized. This observation strengthens our argument for modeling personalized affective responses rather than pursuing consensus annotations.
\newpage
\begin{figure*}[!htbp]
    \centering
    \includegraphics[width=1\linewidth]{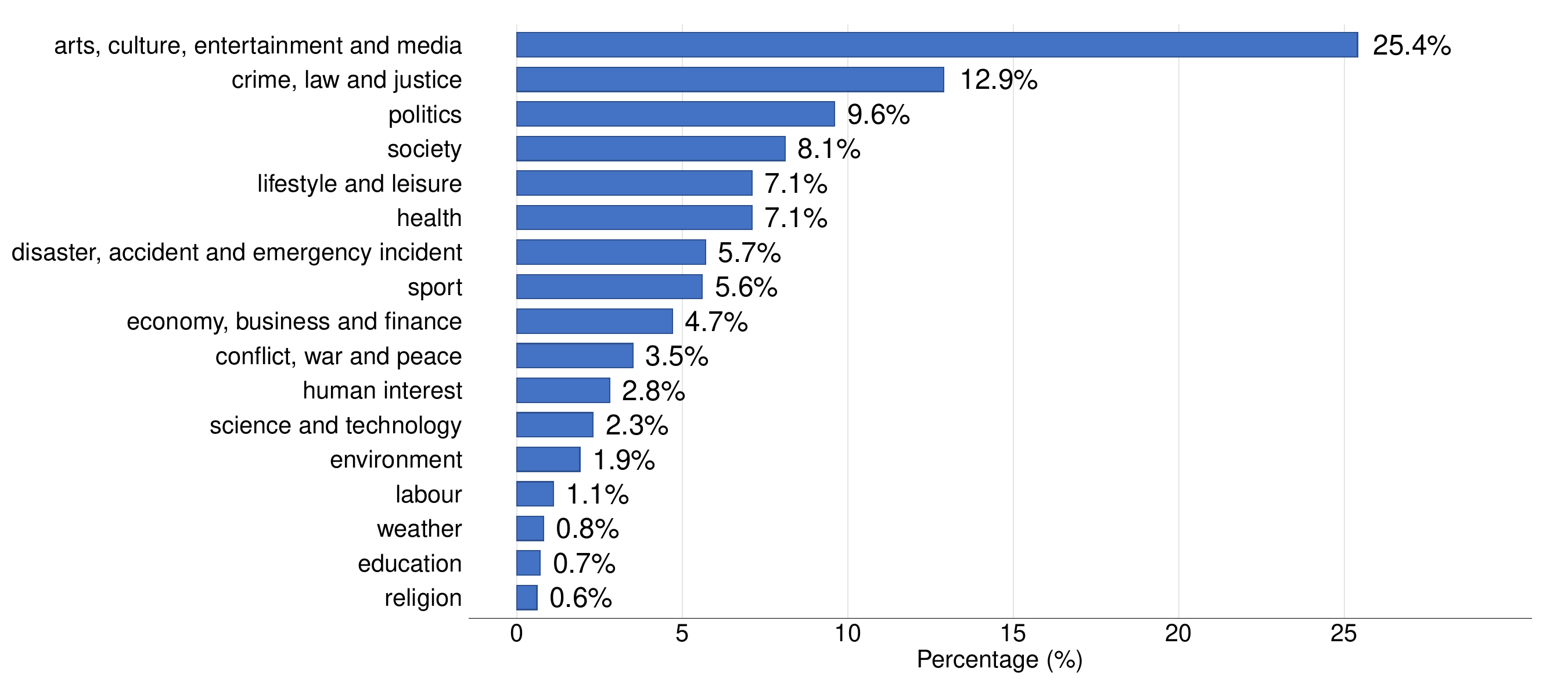}
    \caption{Distribution of news articles across topic categories.}

    \label{fig:topic_distribution}
\end{figure*}

\begin{figure*}[!htbp]
    \centering
    \includegraphics[width=0.82\linewidth]{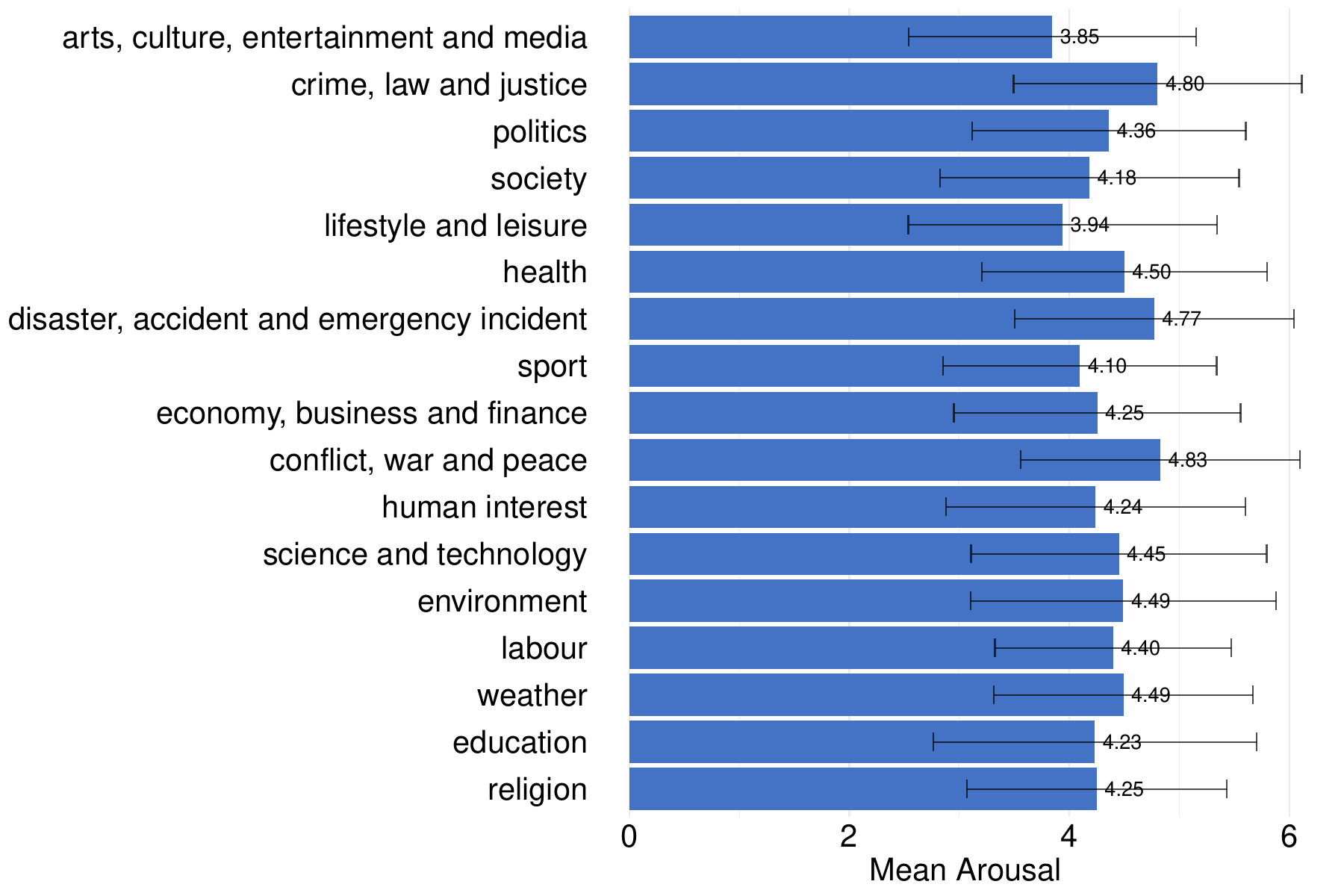}
    \caption{Average arousal scores by topic category, with error bars indicating one standard deviation from the mean.}

    \label{fig:mean_arousal_per_topic}
\end{figure*}

\subsection{Topic Classification Details}
We apply the following prompt in JSON mode with the gemini-1.5-pro endpoint. 

\label{sec:classification_prompt}
\begin{csquote}
\begin{Verbatim}[breaklines=true]
You are an expert news content analyst. 

**Task 1:**

Your task is to describe the IMAGE in this Facebook news post and how it works together with the post's text. Focus on:

- What's shown in the image
- How the image and text complement or contrast with each other
- Any visual elements that particularly grab attention (e.g., graphics, colors, composition)

**Task 2:**

Focus only on the attached the IMAGE of the news post to select the SINGLE most appropriate category:

Categories and definitions

**Task 3:**

Provide a confidence score (1-100) indicating how certain you are of your choice in Task 2. 

**Task 4:**

Analyze both the text content and image to select the SINGLE most appropriate category and provide a confidence score (1-100):

Categories and definitions

- arts, culture, entertainment and media: All forms of arts, entertainment, cultural heritage and media
- conflict, war and peace: Acts of socially or politically motivated protest or violence, military activities, geopolitical conflicts, as well as resolution efforts
- crime, law and justice: The establishment and/or statement of the rules of behaviour in society, the enforcement of these rules, breaches of the rules, the punishment of offenders and the organisations and bodies involved in these activities
- disaster, accident and emergency incident: Man made or natural event resulting in loss of life or injury to living creatures and/or damage to inanimate objects or property
- economy, business and finance: All matters concerning the planning, production and exchange of wealth.
- education: All aspects of furthering knowledge, formally or informally
- environment: All aspects of protection, damage, and condition of the ecosystem of the planet earth and its surroundings.
- health: All aspects of physical and mental well-being
- human interest: Item that discusses individuals, groups, animals, plants or other objects in an emotional way
- labour: Social aspects, organisations, rules and conditions affecting the employment of human effort for the generation of wealth or provision of services and the economic support of the unemployed.
- lifestyle and leisure: Activities undertaken for pleasure, relaxation or recreation outside paid employment, including eating and travel.
- politics: Local, regional, national and international exercise of power, or struggle for power, and the relationships between governing bodies and states.
- religion: Belief systems, institutions and people who provide moral guidance to followers
- science and technology: All aspects pertaining to human understanding of, as well as methodical study and research of natural, formal and social sciences, such as astronomy, linguistics or economics
- society: The concerns, issues, affairs and institutions relevant to human social interactions, problems and welfare, such as poverty, human rights and family planning
- sport: Competitive activity or skill that involves physical and/or mental effort and organisations and bodies involved in these activities
- weather: The study, prediction and reporting of meteorological phenomena

**Task 5:**

Provide a confidence score (1-100) indicating how certain you are of your choice in Task 4.
\end{Verbatim}
\end{csquote}

\subsection{Regression Analysis Details}
This section provides additional details on the regression models used in the main text (Section~\ref{sec:regression_analysis}), including full model specifications, results for additional models, and a discussion of variable importance.

\paragraph{Model Specifications and Estimation}
We employ linear mixed-effects models (LMMs) to analyze the influence of persona variables and other factors on annotators' arousal ratings. LMMs are appropriate for this analysis because they account for the nested structure of the data (multiple annotations per news post and per annotator) and allow for both fixed effects (e.g., persona variables) and random effects (e.g., individual differences between annotators and news posts).
All models are estimated using the \texttt{lme4} package ~\cite{JSSv067i01} in R. The dependent variable in all models is the annotator's arousal rating for a given news post (ranging from 1 to 7).

The following models are estimated in the main text in Section~\ref{sec:regression_analysis}:

\begin{enumerate}
    \item \textbf{Null Model:} Baseline model with only a random intercept for news text.
    \begin{Verbatim}[breaklines=true]
    Arousal ~ 1 + (1 | Text)
    \end{Verbatim}

    \item \textbf{Persona Model:} Includes 47 persona variables as fixed effects and a random intercept for news text.
    \begin{Verbatim}[breaklines=true]
    Arousal ~ PersonaVariables + (1 | Text)
    \end{Verbatim}
    where \texttt{PersonaVariables} represents the full set of 47 persona variables.

    \item \textbf{User Model:} Includes random intercepts for both news text and annotator ID.
    \begin{Verbatim}[breaklines=true]
    Arousal ~ 1 + (1 | Text) + (1 | UserID)
    \end{Verbatim}
\end{enumerate}

\paragraph{Additional Models}

To explore the contributions of other contexual factors, we estimate these additional models:

\begin{enumerate}
    \setcounter{enumi}{3}
    \item \textbf{Outlet Model:} Adds news outlet as a fixed effect.
    \begin{Verbatim}[breaklines=true]
        Arousal ~ PersonaVariables + Outlet + (1 | Text)
        \end{Verbatim}

    \item \textbf{Calibration Model:} Adds responses to the three calibration items as fixed effects.
    \begin{Verbatim}[breaklines=true]
    Arousal ~ PersonaVariables + Calibration + (1 | Text)
    \end{Verbatim}

    \item \textbf{Topic Model:} Adds news post topic category as a fixed effect.
    \begin{Verbatim}[breaklines=true]
    Arousal ~ PersonaVariables + Topic + (1 | Text)
    \end{Verbatim}

    \item \textbf{All Model:} Combines all fixed effects from the Outlet, Calibration, and Topic models.
    \begin{Verbatim}[breaklines=true]
    Arousal ~ PersonaVariables + Outlet + Calibration + Topic + (1 | Text)
    \end{Verbatim}

    \item \textbf{All + User Model:} Adds a random intercept for user ID to the All Model.
    \begin{Verbatim}[breaklines=true]
     Arousal ~ PersonaVariables + Outlet + Calibration + Topic + (1 | Text) + (1|UserID)
    \end{Verbatim}
\end{enumerate}

\paragraph{Full Regression Results}
Table~\ref{tab:regression-results-full} presents the full results for all models, including marginal and conditional $R^2$ values, calculated using the method described by~\citet{https://doi.org/10.1111/j.2041-210x.2012.00261.x}.

\begin{table*}[t]
\centering
\begin{tabular}{l@{\hspace{12pt}}l@{\hspace{12pt}}l@{\hspace{12pt}}cc}
\toprule
& & & \multicolumn{2}{c}{\textbf{Variance Explained}} \\
\cmidrule(lr){4-5}
\textbf{Model} & \textbf{Fixed Effects} & \textbf{Random Effects} & \textbf{Marginal $R^2$} & \textbf{Conditional $R^2$} \\
\midrule
\multicolumn{5}{l}{\textit{Baseline Models}} \\
Null & None & Text & 0.000 & 0.131 \\
User & None & Text + User & 0.000 & 0.317 \\[6pt]

\multicolumn{5}{l}{\textit{Individual Differences}} \\
Persona & Persona & Text & 0.152 & 0.286 \\
Persona + User & Persona & Text + User & 0.139 & 0.377 \\[6pt]

\multicolumn{5}{l}{\textit{Content \& Behavior}} \\
Outlet & Persona + Outlet & Text & 0.166 & 0.287 \\
Calibration & Persona + Calibration & Text & 0.193 & 0.328 \\
Topic & Persona + Topics & Text & 0.217 & 0.283 \\[6pt]

\multicolumn{5}{l}{\textit{Combined Models}} \\
All & Persona + All Above & Text & \textbf{0.261} & 0.326 \\
All + User & Persona + All Above & Text + User & 0.239 & \textbf{0.383} \\
\bottomrule
\end{tabular}
\caption{Regression analysis of affective arousal to news headlines. Models progress from baseline through increasingly complex specifications, incorporating individual differences (persona variables), content features (outlet, topic), and behavioral measures (calibration). Marginal $R^2$ shows variance explained by fixed effects alone, while conditional $R^2$ includes both fixed and random effects.}
\label{tab:regression-results-full}
\end{table*}

\paragraph{Variable Importance}
\label{sec:variable_importance}
What are the most important persona variables? Is it more the case that some specific persona variables explain the vast majority of variance or is it rather spread out across all variables? To answer this question, we analyze the Persona model and calculated the Eta-squared ($\eta^2$), a commonly used measure representing the proportion of the total variance in the dependent variable accounted for by a given independent variable. The calculations are performed using the \texttt{effectsize} package \citep{effectsize} in R.

\begin{table}[!h]
\centering
\begin{tabular}{l r}
\toprule
Parameter & Partial $\eta^2$ \\
\midrule
Personal Income (GBP) & 0.010 \\
Agreeableness & 0.009 \\
Neuroticism & 0.008 \\
Employment Status & 0.008 \\
Television & 0.008 \\
Current Emotional State (PANAS) & 0.008 \\
Highest Education Level Completed & 0.008 \\
News Trust: Independent & 0.007 \\
News Trust: Mirror & 0.007 \\
Extraversion & 0.007 \\
\bottomrule
\end{tabular}
\caption{Top 10 variables with the largest effect sizes (partial $\eta^2$) in the Persona Model}
\label{tab:effect_size}
\end{table}

Based on effect sizes, the individual contributions of the persona variables to explaining variance in arousal are generally modest. The majority of persona variables have small effect sizes, below 0.005. We show the top 10 persona variable with highest effect sizes in Table~\ref{tab:effect_size}. 
Despite this overall trend, a subset of variables exhibit somewhat larger effect sizes. These included factors related to socioeconomic status, such as personal income ($\eta^2 = 0.010$) and education level ($\eta^2 = 0.008$), as well as employment status ($\eta^2 = 0.008$). 
Personality traits also demonstrate notable influence, particularly Agreeableness ($\eta^2 = 0.009$) and Neuroticism ($\eta^2 = 0.008$). Among media consumption patterns, television viewing habits stand out ($\eta^2 = 0.008$), while current emotional state also show meaningful effects ($\eta^2 = 0.007$).

These findings suggest that while the regression model as a whole demonstrates a reasonable ability to predict arousal (as indicated by the R$^2$ values discussed previously), the influence of individual persona variables is, for the most part, limited. The observed model fit likely stems from the cumulative effect of numerous variables with small individual contributions. This pattern aligns with the complex, multifaceted nature of affective responses to news content, where multiple personal characteristics interact to shape individual reactions (also see the quantitative interview analysis in Section \ref{sec:qual_interview}. The distributed nature of these effects underscores the importance of considering a broad spectrum of persona variables in modeling affective responses, rather than focusing on a limited set of characteristics.

\paragraph{Analysis of Content and Behavioral Effects}
While our analyses in the main paper focus on modeling individual differences through user-level variables, our dataset contains rich metadata about the content itself: news topics, headline image categories (see Section~\ref{sec:classification_prompt}), and source outlets. We also collected calibration data by having annotators respond to three standardized items from the ANET dataset. To understand the relative importance of these factors, we first evaluate their contributions separately (Outlet, Calibration, and Topic models) before combining them (All model).

The Outlet and Topic models, which incorporate static content features, achieve similar total explanatory power (conditional $R^2$) to the Persona model but with higher fixed-effect contributions (marginal $R^2$). This suggests these content-based features capture some of the variance previously attributed to random effects, without improving overall prediction. In contrast, the Calibration model shows higher total explanatory power ($R^2 = 0.328$ vs. $0.286$), indicating that annotators' annotation behavior on the calibration items may potentially capture variance unexplained by our carefully selected persona variables.

The All model, despite incorporating numerous fixed effects, maintains approximately the same conditional $R^2$ as the Calibration model. However, it demonstrates a substantial shift in $R^2$ distribution, with marginal $R^2$ reaching 0.261—exceptional for annotator modeling in NLP~\cite{hu-collier-2024-quantifying}. Notably, we achieve better conditional $R^2$ compared to the User model (random-effects only), which only includes random intercepts for text stimuli. This improvement likely stems from two factors: first, the inherently conservative nature of random-effects fitting, which employs regularization to prevent overfitting; and second, random effects' limitation in capturing structural information within the data. While random effects excel at modeling individual-level variation, they treat such variation as purely stochastic, potentially overlooking systematic patterns that our comprehensive set of fixed effects can capture. Our results demonstrate that affective responses to news content, though complex, exhibit structural patterns that can be systematically modeled through carefully selected persona variables, including demographic characteristics, psychological traits, and news consumption behaviors.

\paragraph{Analysis of User Random Effects}
Given the previous results, we then investigate whether adding user-level random effects benefits models with rich fixed effects. In theory, perfect fixed effects would eliminate the need for user-level random effects. In practice, however, adding user-level random effects improves model fit for both the Persona (Personal + User model) and All (All + User model) models, though with diminishing returns. The improvement is smaller for the All model ($\Delta = 0.057$) compared to the Persona model ($\Delta = 0.091$), suggesting we may be approaching a ceiling for random effects gains. This asymptotic behavior indicates that while better fixed effects reduce the potential contribution of random effects, our current setup has not yet exhausted all relevant fixed effect variables, leaving room for future data collection and modeling improvements.

\paragraph{Discussion of Regression Results}
Our findings connect to a fundamental question in psychology: do people's reactions come from who they are (their personality, beliefs, demographics) or from what they're responding to (in our case, the news content)? Our results indicate both, supporting an interactionist perspective~\cite{Mischel1995} - person-level variables and stimulus (the news posts) both contribute meaningfully to explaining affective responses, with their combination yielding higher explanatory power. 

The persistent benefit of including user-level random effects, even in our most comprehensive model ($\Delta R^2 = 0.057$
), aligns with contemporary personality theory~\cite{FLEESON201582} which conceptualizes individual differences through density distributions. This framework suggests that while considerable behavioral variability exists within each individual, the parameters of these distributions may be stable across. In our case, this means that while a person's affective responses to news may vary substantially across different stories, their pattern of variation itself could be characteristic and predictable. This theoretical perspective helps explain why both fixed effects (capturing systematic individual differences) and random effects (accounting for person-specific response patterns) contribute uniquely to our model's predictive power.

\subsection{Post-Annotation Questionnaire}
\label{sec:qual_interview}

To better understand how annotators approached the task and complement the quantitative analysis of persona variables, we conduct a post-annotation qualitative study using a detailed questionnaire. The questionnaire is shown below. Following the questionnaire, we present an in-depth analysis of the responses for each question.

\begin{csquote}
\interviewq{1}{How do you think your personal background (e.g., age, education, political views) influenced your emotional responses to the news headlines?}
\interviewq{2}{Did you notice any patterns in the types of headlines that elicited stronger emotional responses from you? If so, what were they?}
\interviewq{3}{How do you think your emotional responses to these headlines might differ from those of the general public?}
\interviewq{4}{How do you think your media consumption habits (e.g., frequency, preferred sources) might have affected your responses to these Facebook news posts? Did you notice any differences in your responses to news posts from different sources or publishers?}
\interviewq{5}{Reflecting on your experience annotating these news posts, what do you believe were the top 3-5 factors that most influenced your emotional responses? These could be related to the news content itself, your personal background, or external circumstances. For each factor, please briefly explain how you think it affected your reactions.}
\interviewq{6}{Reflecting on your experience with this annotation task and how you typically consume news, are there any insights, observations, or personal reflections you'd like to share about how you engaged with and responded to the news posts, or anything else you'd like to share?}
\end{csquote}

\subsubsection{Q1}
Regarding the influence of personal background (Q1), annotators demonstrate a keen awareness of how factors including age and lived experiences, political affiliations, educational background, media literacy and consumption habits and personal values shape their emotional processing of news. For instance, one annotator reflects on how their generation's experience during the cold war impacts their reactions to current events, stating that they ``get this pit in my stomach when I read these stories'' due to specific events experienced during their lifetime, which differs from the experiences of younger people. Another annotator emphasizes the impact of political views on their emotional responses, noting that they feel ``really frustrated and annoyed'' towards content that conflicted with their political ideology. These examples illustrate how personal history and deeply held beliefs create unique perspectives and biases, coloring readers' emotional engagement with the news. Additionally, many annotators report becoming desensitized due to constant exposure to negative news and recognized modern phenomena like clickbait. There is a notable awareness of how different news sources operate, with some annotators expressing inherent distrust of certain outlets.

\subsubsection{Q2}
\label{sec:open_ended_Q2}
When analyzing the types of headlines that elicit stronger responses (RQ2), we observe a clear distinction between content-driven and presentation-driven factors. Regarding content, annotators consistently identify news related to harm, suffering, and threats to vulnerable populations as powerful emotional triggers. One annotator's comment captures this pattern: ``I really feel it more when the story is about people getting hurt, especially when it's kids or families.'' Contemporary societal issues also generate intense responses, with annotators citing topics such as COVID-19, immigration, healthcare systems, and international conflicts. Personal relevance emerges as another crucial content factor, with annotators responding more intensely to news that mirrors their experiences or aligns with their values. As one annotator puts it: ``when it's something I've been through myself, or it reminds me of my own family, it really gets to me.''

In terms of presentation, visual elements significantly influence emotional intensity. Multiple annotators report that headlines accompanied by images, especially those depicting suffering or tragedy, elicit stronger emotional reactions, with some finding certain visual content overwhelming. This finding validates our research design's inclusion of complete news post screenshots rather than headlines alone. Source credibility also shapes emotional engagement, with annotators expressing greater trust in established news sources (e.g., BBC) compared to social media, and demonstrating skepticism toward tabloids and sensationalized content.

\subsubsection{Q3}

In exploring potential differences between their responses and those of the ``general public'' (Q3), responses mention both universality and divergence. While there is acknowledgment of shared emotional ground, particularly regarding responses to tragedy, suffering, and social norm violations, these mentions are often qualified by extensive discussion of individual variations. A strong theme emerges around the recognition of response variability, with one annotator articulating that "everyone's got their own way of feeling about things - you can't expect two people to react exactly the same." Participants frequently discuss how their personal characteristics - including educational background, socioeconomic status, professional experience, and neurodiversity - shape their responses. Many believe their reactions deviate from the perceived norm, either describing themselves as more analytical compared to a generally more ``empathetic'' public, or reporting stronger emotional engagement than average. Notably, several participants challenge the very concept of a ``general public,'' emphasizing the diversity of perspectives and questioning such generalizations, with one observing that readers of certain newspapers are ``conditioned'' to react with greater anger to headlines. 
\subsubsection{Q4}

When asked about the influence of their media consumption habits (Q4), an interesting disconnect emerges. Many annotators explicitly state that their media consumption patterns do not affect their responses, yet their explanations reveal deep-seated attitudes toward different news sources. This apparent contradiction stems from annotators viewing their skepticism toward certain platforms and outlets not as a ``consumption pattern'' but as a fundamental approach to information processing. Many annotators express a high degree of distrust towards social media platforms such as Facebook as a primary news source and towards tabloid outlets, contrasting these with more trusted, traditional sources like the BBC. One annotator, highlighting their distrust of certain outlets, states that they avoid tabloids because they are ``just nonsense really, proper biased'' while another express a general suspicion of Facebook posts, viewing the platform as more for social interaction than trustworthy news. 
However, they do not view these preferences as biasing their responses, but rather as applying consistent critical evaluation. Additionally, annotators broadly fall into two groups regarding their approach to source evaluation. The first group reports that source credibility significantly influence their emotional engagement, with one noting they ``don't get as worked up about stories from dodgy sources.'' The second group emphasizes prioritizing content over source, with one explaining they ``only consider the content, not the publisher.''

\subsubsection{Q5}

Reflecting on the most salient factors shaping their emotional responses (Q5), annotators frequently emphasize an interplay of several key elements. Personal background such as upbringing and professional experience emerge as particularly important. Similar numbers of annotations mention news content, with annotators particularly responsive to stories involving injustice, vulnerable populations (especially children), and issues of immediate personal relevance. One annotator powerfully illustrate this interaction between these two factors, explaining how ``growing up working class'' might have instilled a certain resilience, yet emphasizing that this does not diminish their emotional response to the suffering of innocent individuals, especially children. The perceived credibility of news sources is also a factor, with one annotator articulating how ``I take proper news sources more seriously.'' The presentation style of news content - including emotional language, imagery, and formatting - also influence responses, with several annotators demonstrating awareness of ``sensationalised'' content and clickbait tactics. Notably, these factors often operate interactively rather than in isolation.

\subsubsection{Q6}

Finally, when considering their experience with the annotation task itself (Q6), many annotators report that the task heighten their awareness of emotional responses to news. Some note that the annotation process make them more consciously aware of their emotional responses, prompting deeper reflection on the quality and factual nature of the news. As one annotator puts it, "I found myself properly thinking about how each story affected me." Participants frequently discuss their news evaluation strategies, considering multiple factors including source credibility, visual elements, and headline framing. Notably, contrary to common assumptions about social media engagement, many annotators express reluctance to share news on social platforms, with one annotator stating that they ``don't share news on social media at all.''

The responses reveal important individual differences in emotional engagement with news. Some annotators, particularly those identifying as neurodivergent, describe carefully managing their emotional engagement to avoid exhaustion, noting that news stories can trigger intense, lasting emotional responses. Others report preferring to reserve emotional energy for personal relationships rather than news content. Content preferences emerge as another key theme, with participants expressing greater interest in positive news, scientific developments, and locally relevant stories, while showing less engagement with celebrity news or sensationalized content.

\subsubsection{Methodological Limitation}
While our qualitative analysis provides valuable insights, several methodological limitations warrant discussion. First, resource constraints necessitate written questionnaires rather than in-depth interviews, potentially limiting the nuance and richness of responses. Second, the opt-in nature of the interview participation may introduce selection bias, as annotators willing to provide detailed written responses might not fully represent our broader annotator population. Third, since the interviews are conducted after both the main annotation task and the persona variable survey, participants' responses might have been influenced by these prior experiences.

These design elements ultimately strengthen rather than compromise our findings. The post-task timing of the interviews prove advantageous, allowing annotators to develop more nuanced reflections on their annotation process and emotional responses. While our participants may represent more engaged annotators, their detailed accounts provide exactly the kind of rich, experiential data needed to complement our structured persona variables. The qualitative insights thus serve their intended purpose: providing crucial context that enriches our understanding of the quantitative patterns observed in our regression analysis.

\subsection{Example Image Textual Description}
\label{sec:example_texual_description}

We show an example news post as well as the textual description generated with the prompt in \ref{sec:classification_prompt}.

\begin{figure}[!h] %
    \centering
    \includegraphics[width=1\linewidth]{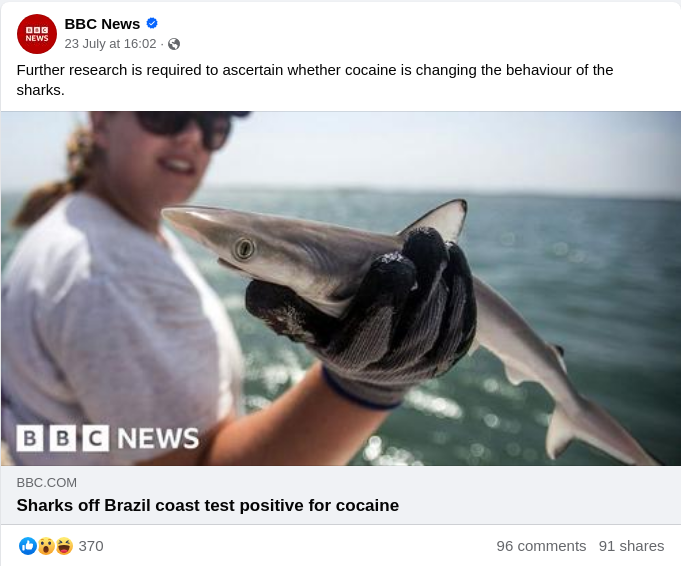} %
    \caption{An example news headline.}
    \label{fig:sample} %
\end{figure}

\begin{csquote}
The image shows a small shark being held by a person wearing gloves. The shark is facing the camera and appears to be alive. The person holding the shark is slightly out of focus, with only the lower part of their face and body visible. The background is a blurry expanse of ocean. The BBC News logo is overlaid on the bottom left corner of the image.
The text and image work together to tell the story of sharks testing positive for cocaine off the coast of Brazil. The image provides a visual representation of the subject matter, while the text gives context and further information. The visual element that grabs attention is the shark itself. Its size and proximity to the camera create a sense of immediacy and make the story feel more tangible.
The contrast between the small, seemingly harmless shark and the serious implications of it testing positive for cocaine adds a layer of complexity to the image.
\end{csquote}

\subsection{Predicting Individual Affective Arousal Prompt}
\label{sec:benchmark_prompt}
For textual input:

\noindent System prompt (for the condition with persona); 
\begin{csquote}
\begin{Verbatim}[breaklines=true]
Today is {current_date}. You are from the United Kingdom. Your first language is English. Here are how you answered a list of questions about yourself:
Question: {persona_question}
Your Answer: {persona_answer}
......
\end{Verbatim}
\end{csquote}

\noindent User prompt:
\begin{csquote}
\begin{Verbatim}[breaklines=true]
headline_input = A Facebook post from {page_name} posted on {post_date}:
{headline_text} The post contains an image where: {texual_description}

Article headline: {headline}

Engagement metrics:
• Emoji/reaction count: {total_interaction}
• Comments: {comments}
• Shares: {shares}

The arousal scale ranges from very calm (1) to very active (7). At the calm end of this scale (1), you feel completely relaxed, calm, sluggish, dull, sleepy, or unaroused. If you feel completely calm, indicate this by choosing 1 (very calm). At the active end of the scale (7), you are stimulated, excited, frenzied, jittery, wide-awake, or aroused. If you feel completely aroused, choose 7 (very active). If you are not at all excited nor at all calm, choose 4 (neutral). Choose in-between options to indicate intermediate levels of excitement or calmness.

How calm vs. active do you feel after reading this news headline? (Arousal)
1 (very calm)
2 (calm)
3 (somewhat calm)
4 (neutral)
5 (somewhat active)
6 (active)
7 (very active)
Please respond with a single number from 1-7.

If few-shot:
    messages.append("role": "user", "content": {headline_input}) 
    messages.append({"role": "assistant", "content": {label})
messages.append({"role": "user", "content": {headline_input}) 
\end{Verbatim}
\end{csquote}

For image-input condition, the prompt is the same except that the post-specific textual description is replaced by the news post screenshot.

\subsection{Additional Few-shot learning results}
\begin{figure*}[h]
    \centering
    \begin{subfigure}[b]{\linewidth}
        \centering
        \includegraphics[width=0.6\linewidth]{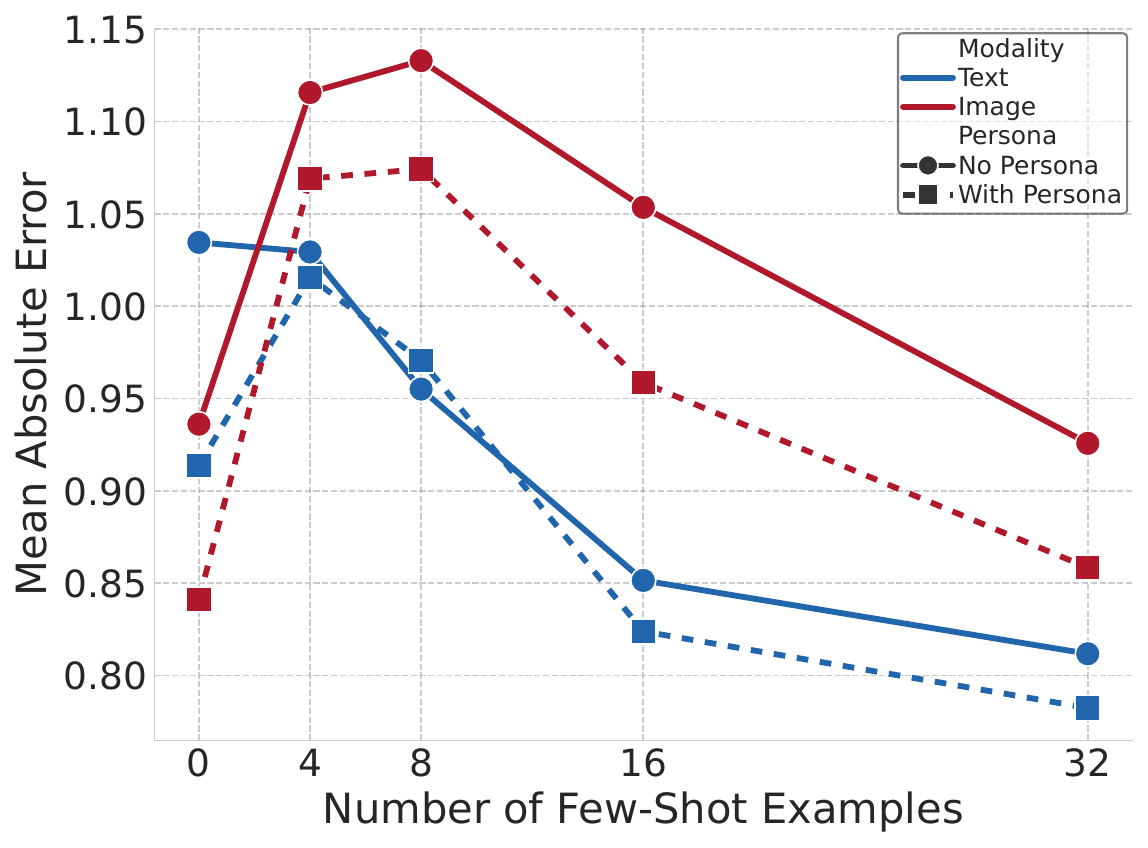}
        \label{fig:subfig1}
    \end{subfigure}
    
    \vspace{1em}  %
    
    \begin{subfigure}[b]{\linewidth}
        \centering
        \includegraphics[width=0.6\linewidth]{images/few_shot_result_exact_accuracy.pdf}
        \label{fig:subfig2}
    \end{subfigure}
    
    \vspace{1em}  %
    
    \begin{subfigure}[b]{\linewidth}
        \centering
        \includegraphics[width=0.6\linewidth]{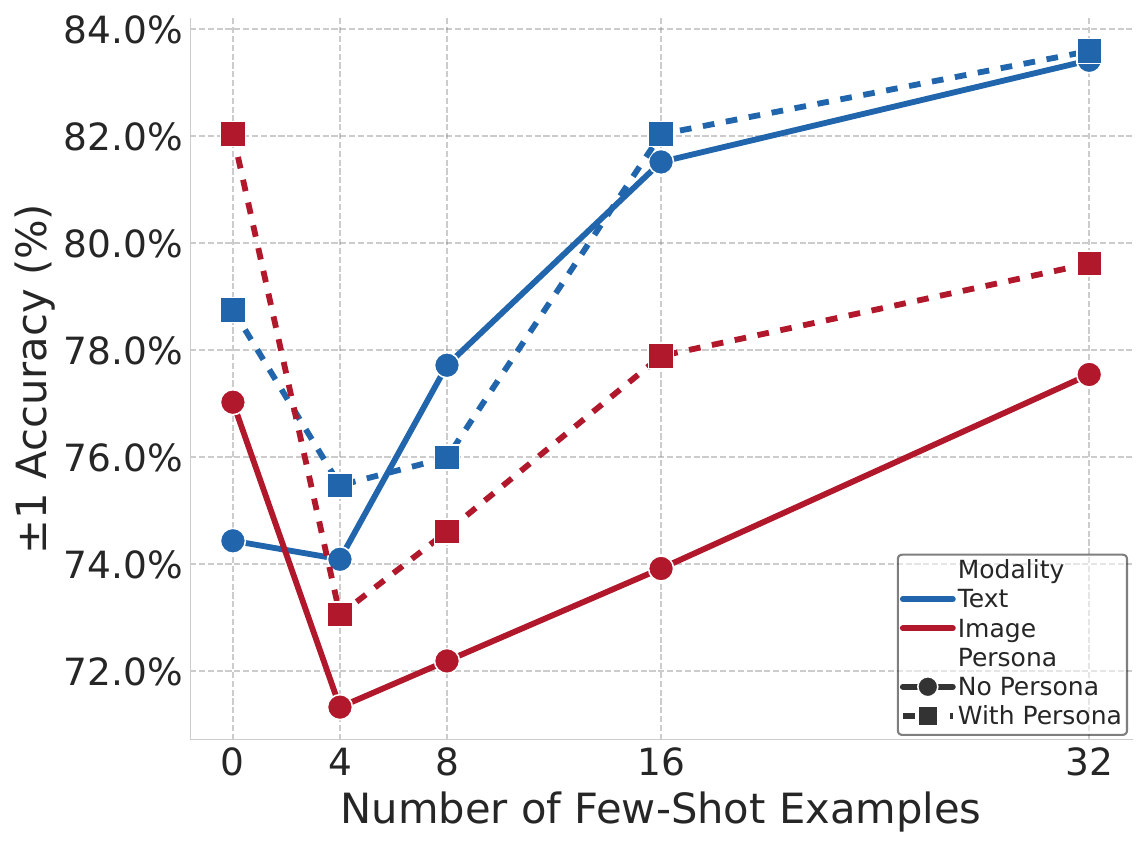}
        \label{fig:subfig3}
    \end{subfigure}
    \caption{Few-shot learning performance, measured by MAE, exact match accuracy (\%), and ±1 accuracy (\%), as a function of the number of few-shot examples (0, 4, 8, 16, and 32).}
    \label{fig:all_subfigs_few_shot}
\end{figure*}

We present additional few-shot results in Table~\ref{tab:few_shot_full_results} and Figure~\ref{fig:all_subfigs_few_shot}.

\begin{table*}[t]
\centering
\begin{tabular}{llrrr}
\toprule
\multirow{2}{*}{\textbf{Modality}} & \multirow{2}{*}{\textbf{Setting}} & \multicolumn{3}{c}{\textbf{Performance Metrics}} \\
\cmidrule(lr){3-5}
& & \textbf{MAE} $\downarrow$ & \textbf{Accuracy} $\uparrow$ & \textbf{Within±1 Accuracy} $\uparrow$ \\
\midrule
\multirow{10}{*}{Text}
 & 0-shot, no persona & 1.035 & 29.4 & 74.4 \\
 & 0-shot, with persona & \textbf{0.914} & \textbf{36.4} & \textbf{78.8} \\
 \cmidrule(lr){2-5}
 & 4-shot, no persona & 1.029 & 31.6 & 74.1 \\
 & 4-shot, with persona & 1.016 & 32.1 & 75.5 \\
 & 8-shot, no persona & 0.955 & 35.1 & 77.7 \\
 & 8-shot, with persona & 0.971 & 36.1 & 76.0 \\
 & 16-shot, no persona & 0.851 & 39.7 & 81.5 \\
 & 16-shot, with persona & 0.824 & 42.3 & 82.0 \\
 & 32-shot, no persona & 0.812 & 42.1 & 83.4 \\
 & 32-shot, with persona & \underline{0.782} & \underline{44.4} & \underline{83.6} \\
\midrule
\multirow{10}{*}{Image}
 & 0-shot, no persona & 0.936 & 37.0 & 77.0 \\
 & 0-shot, with persona & \textbf{0.841} & \textbf{39.6} & \textbf{82.0} \\
 \cmidrule(lr){2-5}
 & 4-shot, no persona & 1.116 & 29.9 & 71.3 \\
 & 4-shot, with persona & 1.069 & 30.6 & 73.1 \\
 & 8-shot, no persona & 1.133 & 27.8 & 72.2 \\
 & 8-shot, with persona & 1.074 & 29.4 & 74.6 \\
 & 16-shot, no persona & 1.054 & 33.3 & 73.9 \\
 & 16-shot, with persona & 0.959 & 35.9 & 77.9 \\
 & 32-shot, no persona & 0.926 & 39.2 & 77.5 \\
 & 32-shot, with persona & \underline{0.858} & \underline{42.8} & \underline{79.6} \\
\bottomrule
\end{tabular}
\caption{Performance comparison of text-based and image-based models across few-shot settings and persona conditions. Lower MAE ($\downarrow$) and higher Accuracy and Within±1 Accuracy($\uparrow$) indicate better performance. Best zero-shot results are in \textbf{bold}, and best overall results per modality are \underline{underlined}. Accuracy and Within±1 accuracy are shown as percentages.}
\label{tab:few_shot_full_results}
\end{table*}

\end{document}